\documentclass{./include/latex-class-tlp/new_tlp}

\usepackage[utf8]{inputenc}
\usepackage{amsmath}
\usepackage{amssymb}
\usepackage{microtype}
\usepackage{url}
\usepackage[]{xcolor} \newcommand{\cellcolor}[1]{\textcolor{red}{#1}}
\usepackage{graphicx}
\usepackage{hyperref}
\usepackage{listings}

\usepackage{tikz}

\makeatletter
\let\@ORGmakecaption\@makecaption
\long\def\@makecaption#1#2{\@ORGmakecaption{#1}{#2}\vskip\belowcaptionskip\relax}
\makeatother

\providecommand{\sysfont}{\textit}

\newcommand{\clingo}{\sysfont{clingo}}
\newcommand{\clingodl}{\clingoM{dl}}

\newcommand{\clingoM}[1]{\clingo{\small\textnormal{[}\textsc{#1}\textnormal{]}}}

\providecommand{\Underscore}{\textunderscore}

\lstdefinelanguage{clingo}{basicstyle=\ttfamily,keywordstyle=[1]\bfseries,keywordstyle=[2]\bfseries,keywordstyle=[3]\bfseries,showstringspaces=false,literate={_}{\Underscore}1 {\%\%}{}0,escapeinside={\#(}{\#)},alsoletter={\#,\&},keywords=[1]{not,from,import,def,if,else,elif,return,while,break,and,or,for,in,del,and,class,with,as,is,yield,async},keywords=[2]{\#const,\#show,\#minimize,\#base,\#theory,\#count,\#external,\#program,\#script,\#end,\#heuristic,\#edge,\#project,\#show,\#sum},keywords=[3]{&,&dom,&sum,&diff,&show},morecomment=[l]{\#\ },morecomment=[l]{\%\ },morestring=[b]",stringstyle={\itshape},commentstyle={\color{darkgray}}}

\lstdefinelanguage{python}{basicstyle=\ttfamily,keywordstyle=[1]\bfseries,showstringspaces=false,literate={_}{\Underscore}{1},escapeinside={\#(}{\#)},alsoletter={\#,\&},keywords=[1]{not,from,import,def,if,else,elif,return,while,break,and,or,for,in,del,and,class,with,as,is,yield,async},morecomment=[l]{\#\ },morestring=[b]",stringstyle={\itshape},commentstyle={\color{darkgray}}}

 \sbox0{\ttfamily A}
\edef\mybasewidth{\the\wd0 }
\sbox0{\scriptsize\ttfamily A}
\edef\mybasewidths{\the\wd0 }
\newcommand\ASP{\lstinline[language=clingo,columns=flexible,mathescape]}
\newcommand\LC[2][0pt]{\makebox[\mybasewidth]{\hspace{#1}$#2$}}
\lstdefinelanguage{clingos}{language=clingo,
  columns=flexible,
  basewidth=\mybasewidths,
  basicstyle=\scriptsize\ttfamily
}
\lstdefinelanguage{clingoht}{language=clingo,
  columns=flexible,
  basewidth=\mybasewidth,
  escapeinside=||,
  mathescape,
  float
}

\newtheorem{definition}{Definition}
\newtheorem{proposition}{Proposition}

\newcommand\indegree{\mathrm{deg}_a^-}
\newcommand\outdegree{\mathrm{deg}_a^+}

\newcommand{\precdot}{\prec\mathrel{\mkern-5mu}\mathrel{\cdot}}

\DeclareFontFamily{U}{mathb}{\hyphenchar\font45}
\DeclareFontShape{U}{mathb}{m}{n}{
<-6> mathb5 <6-7> mathb6 <7-8> mathb7
<8-9> mathb8 <9-10> mathb9
<10-12> mathb10 <12-> mathb12
}{}
\DeclareSymbolFont{mathb}{U}{mathb}{m}{n}
\DeclareMathSymbol{\pprec}{\mathrel}{mathb}{"CE}

\newcommand{\events}{\ensuremath{\mathcal{E}}}

\newcommand{\pos}[2]{\ensuremath{v_{#1,#2}}}

\newcommand{\comment}[1]{}
\newcommand{\note}[1]{\comment{\textcolor{blue}{#1}}}
\newcommand{\todo}[1]{\comment{\textcolor{red}{#1}}}
\newcommand{\REWc}[2]{#1}

\title[Routing and Scheduling]{Routing and Scheduling in Answer Set Programming applied to Multi-Agent Path Finding:\\ Preliminary Report}

\author[Jan Behrens et al.]{JAN BEHRENS\\
  University of Potsdam, Germany
  \and
  ROLAND KAMINSKI and TORSTEN SCHAUB\\
  University of Potsdam, Germany and Potassco Solutions, Germany
  \and
  TRAN CAO SON\\
  New Mexico State University, USA
  \and
  JIŘÍ ŠVANCARA\\
  Charles University, Prague, Czech Republic
  \and
  PHILIPP WANKO\\
  Potassco Solutions, Germany
}

\begin{document}

\maketitle

\begin{abstract}
  We present alternative approaches to routing and scheduling in Answer Set Programming (ASP),
  and explore them in the context of Multi-agent Path Finding.
The idea is to capture the flow of time in terms of partial orders rather than time steps attached to actions and fluents.
  This also abolishes the need for fixed upper bounds on the length of plans.
The trade-off for this avoidance is that (parts of) temporal trajectories must be acyclic,
  since multiple occurrences of the same action or fluent cannot be distinguished anymore.
While this approach provides an interesting alternative for modeling routing,
  it is without alternative for scheduling since fine-grained timings cannot be represented in ASP in a feasible way.
This is different for partial orders that can be efficiently handled by external means
  such as acyclicity and difference constraints.
We formally elaborate upon this idea and present several resulting ASP encodings.
  Finally, we demonstrate their effectiveness via an empirical analysis.
\end{abstract}
 \begin{keywords}
  Answer Set Programing,
  Routing,
  Scheduling,
  Multi-Agent Path Finding
\end{keywords}
\section{Introduction}\label{sec:introduction}

The ease of Answer Set Programming (ASP~\cite{lifschitz19a}) to express reachability
has made it a prime candidate for addressing routing problems, such as
multi-agent path finding \cite{erkiozsc13a},
phylogenetic inference \cite{brerermiri07a},
wire routing \cite{erliwo00a},
etc.
This lightness vanishes, however, once routing is combined with scheduling for expressing deadlines and durations
since fine-grained timings cannot be feasibly represented in ASP.
This is because ASP~\cite{lifschitz02a}, just like  CP~\cite{balapanu06a} and SAT~\cite{rintanen09a},
usually account for time by indexing action and fluent variables with time steps,
a technique tracing back to situation calculus~\cite{mcchay69a} and temporal logic~\cite{kamp68a}.
Each time step results in a copy of the problem description.
Hence, the finer the granularity of time, the more copies are produced.
Although this is usually a linear increase (in terms of plan length), it eventually leads to a decrease in performance.

We rather capture flows of time by means of partial orders on actions and/or fluents,
similar to partial-order planning~\cite{sacerdoti75a}.
In fact,
the avoidance of time steps also eliminates the need for upper bounds on temporal trajectories, that is, horizons or makespans.
The trade-off for this is that (parts of) temporal trajectories must be acyclic,
since multiple occurrences of the same action or fluent cannot be distinguished without indexing.
Moreover,
to cease the influence of the granularity of time on the solving process,
the idea is to outsource the treatment of partial orders by using hybrid ASP,
more precisely, acyclicity and difference constraints.
Intuitively, these constraints are used for ordering actions in order to avoid conflicts among them.

As a matter of fact,
we have already applied this technique in several industrial-scale applications involving routing and scheduling,
namely,
train scheduling~\cite{abjoossctowa21a},
system design~\cite{hamunescwa23a}, and
warehouse robotics~\cite{rascwachliso23a}. However,
the intricacy of these applications obscured a clear view on the underlying encoding techniques and their formal foundations,
which we present in what follows.
To simplify this,
we apply our approach to Multi-Agent Path Finding (MAPF~\cite{ststfekomawaliatcokubabo19b}),
a simple yet highly relevant AI problem.

Our paper is organized as follows.
Section~\ref{sec:background} gives some basic concepts and notation
from graph theory and provides a gentle introduction to MAPF.

Section~\ref{sec:routing} is dedicated to collision-free routing.
We start in Section~\ref{sec:ordering}
by introducing an alternative characterization of MAPF based on event orderings.
Each event represents a position of an agent.
As a reference,
we first give in Section~\ref{sec:mapf:encoding:vanilla} an encoding of MAPF
in accordance with the traditional approach of Answer Set Planning~\cite{lifschitz02a,sopobasc23a}.
We then develop a new encoding for MAPF in Section~\ref{sec:mapf:encoding:ordering},
closely following our characterization from Section~\ref{sec:ordering}, and prove soundness and completeness.
The underlying encoding technique
relies on acyclicity constraints and
drops time steps and explicit bounds on the length of plans.

Section~\ref{sec:scheduling} combines routing with scheduling by considering durative actions.
To this end,
we begin with a definition of weighted MAPF, enriching the base case with durations and safety periods.
Also, we lift the concepts of vertex, swap, and follow conflicts to the weighted case.
In analogy to Section~\ref{sec:ordering},
we introduce in Section~\ref{sec:mapping} a characterization of weighted MAPF
by associating arrival times with events.
As before,
we first give in Section~\ref{sec:wmapf:encodings:vanilla} an encoding of weighted MAPF
following traditional Answer Set Planning.
We then develop in Section~\ref{sec:wmapf:encodings:sequence} an encoding for weighted MAPF,
reflecting the characterization from Section~\ref{sec:mapping}, and prove soundness and completeness.
Notably, this encoding is more or less obtained from that in Section~\ref{sec:mapf:encoding:ordering}
by merely replacing acyclicity by difference constraints.
Accordingly,
it also drops time steps and explicit bounds on the length of plans.
As a consequence, fine-grained scheduling is no burden on ASP solving any longer
but rather outsourced to an underlying difference constraints propagator.

\comment{T: possibly expand once sections are filled}
\todo{Correct me!}
Section~\ref{sec:experiments} provides an empirical analysis of the whole spectrum of alternative routing and scheduling encodings.
The underlying experiments are run on a variety of benchmark classes capturing various MAPF scenarios.
\todo{Correct me!}
We discuss related work and summarize our approach in Section~\ref{sec:discussion}.

 \section{Background}\label{sec:background}

We begin by fixing some preliminaries from graph theory~\cite{benwil10a}.
We consider \emph{graphs} $(V,E)$
where $V$ is a finite set of \emph{vertices} and $E \subseteq V \times V$ is a set of \emph{edges}.
A \emph{walk} $\pi$ in a graph $(V,E)$ is a sequence
\(
(v_i)_{i=0}^n
\)
of vertices $v_i\in V$ for $0\leq i\leq n$
such that $(v_i,v_{i+1})\in E$ for all $0 \leq i < n$.
We use $\pi(i) = v_i$ to refer to the vertex at index $0 \leq i \leq n$ of walk $\pi$
and $|\pi| = n$ to refer to the \emph{length} of the walk.
A walk $\pi$ in a graph $(V,E)$ \emph{leads} from $u \in V$ to $v \in V$ if $\pi(0)=u$ and $\pi(n)=v$.
The \emph{vertex set} of a walk $\pi$ in a graph $(V,E)$ is
\(
V(\pi)
=
\{\pi(i)\mid 0\leq i\leq |\pi|\}
\);
we write
\(
v\in \pi
\)
for
\(
v\in V(\pi)
\).
The \emph{index set} of a vertex $v \in V$
in a walk $\pi$ in a graph $(V,E)$ is
\(
\iota_\pi(v)=\{i\mid v=\pi(i), 0\leq i\leq |\pi|\}
\).
A \emph{path} is a walk in which all vertices (and therefore also all edges) are distinct.
A \emph{cycle} is a walk in which only the first and last vertices are equal.
\REWc{Note that a cycle with just one vertex is also a path.}{T: but this is no walk, or\dots?}
A \emph{stroll} in a graph $(V,E)$ is a walk in the reflexively closed graph $(V,E \cup \{ (v,v) \mid v \in V \})$.
That is, a stroll is obtained from a walk in $(V,E)$ by repeating some or none of its vertices
(to mimic waiting).
A stroll $\pi$ is \emph{path-like}, if
\(
\iota_\pi(v)=[\min\iota_\pi(v),\max\iota_\pi(v)]
\)
for all $v \in \pi$.
Informally, a stroll is path-like, if dropping all repeated vertices results in a path.

 We use the above to define the MAPF problem.
In what follows,
we consider \emph{simple} graphs $(V,E)$
where $E\subseteq V\times V$ is irreflexive.
A \emph{MAPF problem} is a triple $(V,E,A)$ where
$(V,E)$ is a finite, simple graph and
$A$ is a finite set of agents.
Each \emph{agent} $a \in A$
has a \emph{start} vertex $s_a \in V$ and
a \emph{goal} vertex $g_a \in V$.
We stipulate that all start and all goal vertices are disjoint.
That is,
we require that $a\neq b$ implies $s_a\neq s_b$ and $g_a\neq g_b$
for all $a,b \in A$.
The start and goal vertex of an agent may coincide,
that is, we may have $s_a=g_a$ for $a\in A$.
An agent can either wait at its current vertex or move to a neighboring one.
Hence, we use strolls to capture the movement of agents.
A \emph{plan} of length~$n$ for a MAPF problem $(V,E,A)$ is a family $\{\pi_a\}_{a\in A}$
of strolls $\pi_a$ of length~$n$ in $(V,E)$
leading from $s_a$ to $g_a$ for all $a \in A$.
A plan for a MAPF problem is \emph{path-based} if all its strolls are path-like.
We use $\iota_a$ as a shortcut for $\iota_{\pi_a}$ whenever it is clear from context that agent $a$ is associated with stroll $\pi_a$.
A plan $\{\pi_a\}_{a \in A}$ of length $n$ for a MAPF problem~$(V,E,A)$ is
\begin{enumerate}
\item \emph{vertex conflict-free}
  if $\pi_a(i) \neq \pi_b(i)$
for all $a,b \in A$ such that $a \neq b$ and $0 \leq i \leq n$,
\item \emph{swap conflict-free}
  if $\pi_a(i+1) \neq \pi_b(i)$ or $\pi_a(i) \neq \pi_b(i+1)$
for all $a,b \in A$ such that $a \neq b$ and $0 \leq i < n$, and
\item \emph{follow conflict-free}
  if $\pi_a(i+1) \neq \pi_b(i)$
for all $a,b \in A$ such that $a \neq b$ and $0 \leq i < n$.
\end{enumerate}
A vertex conflict occurs if two agents occupy the same vertex at some point.
A swap conflict   occurs if two agents traverse the same edge in opposite directions at some point.
A follow conflict occurs if an agent enters a vertex another agent just left.
The absence of follow conflicts implies the absence of swap conflicts.

 Finally,
we rely on a basic acquaintance with ASP, and refer the interested reader for details to the literature~\cite{lifschitz99b};
the input language of \clingo\ is described in the \emph{Potassco User Guide}~\cite{PotasscoUserGuide}.

\section{Routing}
\label{sec:routing}

\begin{figure}[th]
\figrule
\includegraphics{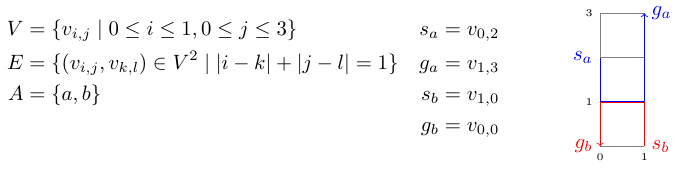}\caption{Example MAPF problem together with stroll candidates}\label{fig:mapf}
\figrule
\end{figure}

\begin{figure}
\figrule
\includegraphics{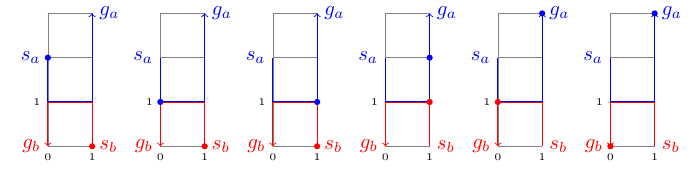}
\caption{Plan $\{\pi_a^5,\pi_b^5\}$ from $\eqref{eq:path:b:fiv}$ without vertex and swap conflicts but with a follow-conflict}
\label{fig:b:swap}
\figrule
\end{figure}

\begin{figure}
\figrule
\centering
\includegraphics{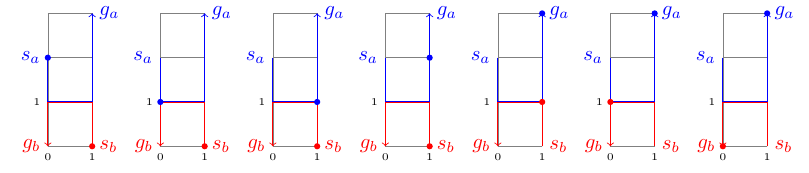}
\caption{Plan $\{\pi_a^6,\pi_b^6\}$  from $\eqref{eq:path:b:six}$ without vertex, swap, and follow conflicts}
\label{fig:b:follow}
\figrule
\end{figure}

Consider the MAPF problem $(V,E,A)$ in Figure~\ref{fig:mapf},
delineating in blue and red the movements of agents $a$ and $b$
all of which traverse vertices $\pos{0}{1}$ and $\pos{1}{1}$
on the following strolls:
\begin{align}
  \label{eq:path:b:for}
  \pi_a^4&= (\pos{0}{2},\pos{0}{1},\pos{1}{1},\pos{1}{2},\pos{1}{3}                      )&
  \pi_b^4&= (\pos{1}{0},\pos{1}{1},\pos{0}{1},\pos{0}{0},\pos{0}{0}                      )\\
  \label{eq:path:b:fiv}
  \pi_a^5&= (\pos{0}{2},\pos{0}{1},\pos{1}{1},\pos{1}{2},\pos{1}{3},\pos{1}{3}           )&
  \pi_b^5&= (\pos{1}{0},\pos{1}{0},\pos{1}{0},\pos{1}{1},\pos{0}{1},\pos{0}{0}           )\\
  \label{eq:path:b:six}
  \pi_a^6&= (\pos{0}{2},\pos{0}{1},\pos{1}{1},\pos{1}{2},\pos{1}{3},\pos{1}{3},\pos{1}{3})&
  \pi_b^6&= (\pos{1}{0},\pos{1}{0},\pos{1}{0},\pos{1}{0},\pos{1}{1},\pos{0}{1},\pos{0}{0})
\end{align}
Each pair of strolls in \eqref{eq:path:b:for} to \eqref{eq:path:b:six} gives rise to a plan:
\begin{itemize}
\item $\{\pi_a^4,\pi_b^4\}$ has length 4 and has a swap conflict, viz.\ $\pi_a(1)=\pi_b(2)$ and $\pi_a(2)=\pi_b(1)$,
\item $\{\pi_a^5,\pi_b^5\}$ has length 5 and has no vertex- and swap-conflicts but a follow conflict, viz.\ $\pi_a(2)=\pi_b(3)$
(its moves are shown in Figure~\ref{fig:b:swap}),
and
\item $\{\pi_a^6,\pi_b^6\}$ has length 6 and is conflict-free
(cf.\ Figure~\ref{fig:b:follow}).
\end{itemize}


\subsection{Event orderings}\label{sec:ordering}

In what follows,
we are interested in characterizing plans for MAPF problems in terms of orders on agent positions.
Since agent positions change over time, we refer to them as \emph{events};
this change is captured by ordering events below.
\begin{definition}
An \emph{event set}~$\events$ for a MAPF problem~$(V,E,A)$ is a set of \emph{events} of form $a@v$ where $a \in A$ and $v \in V$.
\end{definition}

For example,
consider the event sets $\events_a$ and $\events_b$ corresponding to the positions of agents~$a$ and~$b$
in plans $\{\pi_a^5,\pi_b^5\}$ and $\{\pi_a^6,\pi_b^6\}$,
whose strolls are given in~\eqref{eq:path:b:fiv} to~\eqref{eq:path:b:six}, respectively:
\begin{align}
  \events_a&=\{a@\pos{0}{2},a@\pos{0}{1},a@\pos{1}{1},a@\pos{1}{2},a@\pos{1}{3}\}\\
  \events_b&=\{b@\pos{1}{0},b@\pos{1}{1},b@\pos{0}{1},b@\pos{0}{0}\}
\end{align}
Both plans are illustrated in Figures~\ref{fig:b:swap} and~\ref{fig:b:follow},
where events are indicated by solid blue and red discs.
Clearly, the combination $\events_a\cup\events_b$ is also an event set.

\begin{definition}
  An \emph{ordered event set}~$(\events, {\prec})$ for a MAPF problem is a pair of
  events~$\events$ for the problem and a
  relation~$\prec$ establishing a partial order among the events.
\end{definition}
Given an ordered event set~$(\events,{\prec})$,
we use $\precdot$ to denote the \emph{cover} of $\prec$.
That is, $\precdot$ is the smallest relation such that ${\precdot}^*={\prec}$ where $\cdot^*$ is the transitive closure.
The \emph{restriction} of an ordered event set $(\events,{\prec})$ for a MAPF problem $(V,E,A)$ to a single agent $a\in A$
is denoted
$(\events_a,{\prec_a})$
where
$\events_a=\{a@v \in \events\mid v\in V\}$
and
${\prec_a}={\prec} \cap (\events_a\times\events_a)$.

Taking a total order whose cover corresponds to the moves of agents $a$ and $b$ in Figures~\ref{fig:b:swap} and~\ref{fig:b:follow}
results in totally ordered event sets $(\events_a,{\prec_a})$ and $(\events_b,\prec_b)$ for both agents:
\begin{align}
  \label{event:set:a}
  a@\pos{0}{2} &\precdot_a a@\pos{0}{1} \precdot_a a@\pos{1}{1} \precdot_a a@\pos{1}{2} \precdot_a a@\pos{1}{3}\\
  \label{event:set:b}
  b@\pos{1}{0} &\precdot_b b@\pos{1}{1} \precdot_b b@\pos{0}{1} \precdot_b b@\pos{0}{0}
\end{align}
As above, the combination $(\events_a\cup\events_b,{\prec_a}\cup{\prec_b})$ is also an ordered event set.

Next, we make precise when an event order reflects the movement of individual agents.
\begin{definition}\label{def:order:path-based}
  An ordered event set $(\events,{\prec})$ for a MAPF problem~$(V,E,A)$ is \emph{path-based}
  if
  \begin{enumerate}
  \item $(\events_a,{\prec_a})$ is totally ordered with least and greatest elements~$a@s_a$ and~$a@g_a$ for all~$a \in A$, and
  \item $(u,v) \in E$ for all $a@u,a@v \in \events$ with $a@u \precdot_a a@v$.
  \end{enumerate}
\end{definition}

For the MAPF problem $(V,E,A)$ in Figure~\ref{fig:mapf},
the ordered event set $(\events_a\cup\events_b,{\prec_a}\cup{\prec_b})$ is path-based
since it contains the totally ordered event sets $(\events_a,{\prec_a})$ and $(\events_b,{\prec_b})$
with least and greatest elements corresponding to the start and goal positions of agents~$a$ and~$b$
and their covers agreeing with the edges of graph $(V,E)$.

While path-based event sets capture the movement of individual agents,
the following concept accounts for their conflict-free interplay.
\begin{definition}\label{def:order:conflict-free}
  An ordered event set~$(\events,{\prec})$ for a MAPF problem~$(V,E,A)$ is \emph{conflict-free}
  if for all events $a@u,b@u \in \events$ with $a \neq b$ either
  \begin{enumerate}
  \item there exists an event $a@v \in \events$ such that $a@u \precdot_a a@v$ and $a@v \prec b@u$, or
  \item there exists an event $b@v \in \events$ such that $b@u \precdot_b b@v$ and $b@v \prec a@u$.
  \end{enumerate}
\end{definition}

In fact,
the ordered event set $(\events_a\cup\events_b,{\prec_a}\cup{\prec_b})$ above comprises two conflicts
because
\(
{a}@\pos{0}{1},{b}@\pos{0}{1}
\)
as well as
\(
{a}@{\pos{1}{1}},{b}@{\pos{1}{1}}
\)
belong to $\events_a\cup\events_b$ but are not related by ${\prec_a}\cup{\prec_b}$.
The adjacency of $\pos{0}{1}$ and $\pos{1}{1}$ allows us to resolve this by adding
\(
{a}@{\pos{1}{2}}\prec{b}@{\pos{1}{1}}
\)
(since this implies
\(
{a}@{\pos{1}{1}}\prec{b}@{\pos{0}{1}}
\)).
Accordingly,
the ordered event set
\begin{align}\label{ex:event:set:conflict:free}
  (\events_a\cup\events_b,({\prec_a}\cup{\prec_b}\cup\{{a}@{\pos{1}{2}} \prec {b}@{\pos{1}{1}}\})^*)
\end{align}
for the MAPF problem from Figure~\ref{fig:mapf} is (path-based and) conflict-free.

For identifying minimally ordered event sets, we define the following compatibility relation.
\begin{definition}\label{def:order:compatible}
  Two ordered event sets~$(\events,{\prec_1})$ and $(\events,{\prec_2})$
  for a MAPF problem
  are \emph{compatible} if
  \begin{enumerate}
  \item $a@u \prec_1 a@v$ iff $a@u \prec_2 a@v$ for all $a@u,a@v \in \events$, and
  \item $a@u \prec_1 b@u$ iff $a@u \prec_2 b@u$ for all $a@u,b@u \in \events$.
  \end{enumerate}
\end{definition}
In other words,
two ordered event sets are compatible,
if they resolve all conflicts among agents in the same way.

For illustration,
we consider the ordered event set in \eqref{ex:event:set:conflict:free} together with the following two:
\begin{align}
(\events_a \cup \events_b,({\prec_a}\cup{\prec_b}\cup\{a@\pos{1}{3} \prec b@\pos{1}{1}\})^*)\label{eq:comp:b}\\
(\events_a \cup \events_b,({\prec_a}\cup{\prec_b}\cup\{b@\pos{0}{0} \prec a@\pos{0}{1}\})^*)\label{eq:comp:c}
\end{align}
All three ordered event sets are path-based and conflict-free.
However, they differ regarding compatibility.
The ordered event sets in \eqref{ex:event:set:conflict:free} and \eqref{eq:comp:b} are compatible;
agent $b$ is waiting longer in \eqref{eq:comp:b} before starting to move.
The ordered event sets in \eqref{ex:event:set:conflict:free} and \eqref{eq:comp:b} are both incompatible with the one in \eqref{eq:comp:c};
in the former two, agent $a$ starts moving before agent $b$ and in the latter, agent $b$ starts moving before agent $a$.

An ordered event set $(\events,\prec_1)$ is \emph{smaller} than $(\events,\prec_2)$ if
${\prec_1} \subseteq {\prec_2}$.
This allows us to distinguish minimally ordered event sets.
\begin{proposition}\label{stm:order:minimal}
  For each conflict-free path-based ordered event set for a MAPF problem,
  there is a unique minimally compatible conflict-free path-based ordered event set for the problem.
\end{proposition}
We call such an event set a \emph{minimal} conflict-free path-based ordered event set.

The ordered event set in \eqref{ex:event:set:conflict:free} is smaller than the one in \eqref{eq:comp:b}.
In fact, the one in \eqref{ex:event:set:conflict:free} is a minimal conflict-free path-based ordered event set.

As a result,
we obtain the following onto relationship between plans and ordered event sets for MAPF problems.
\begin{proposition}\label{stm:plan:onto:order}
  \begin{enumerate}
    \item\label{stm:plan:onto:order:a}
      For each conflict-free path-based plan for a MAPF problem,
      there exists exactly one minimal conflict-free path-based ordered event set for the problem, and
    \item\label{stm:plan:onto:order:b}
      for each minimal conflict-free path-based ordered event set for a MAPF problem,
      there exists at least one conflict-free path-based plan for the problem.
  \end{enumerate}
\end{proposition}
This onto relationship underlines the role of conflict-free path-based ordered event sets as an abstraction of conflict-free path-based plans;
the essence lies in the order of events, no matter the continuance in a position.
For example,
the conflict-free path-based plan $\{\pi_a^6,\pi_b^6\}$ from \eqref{eq:path:b:six} induces the
minimal conflict-free path-based ordered event set in~\eqref{ex:event:set:conflict:free}
and vice versa.
The same applies to extensions of $\{\pi_a^6,\pi_b^6\}$ repeating vertices,
as long as they preserve the order in~\eqref{ex:event:set:conflict:free}.

Let us now detail the constructions underlying Proposition~\ref{stm:plan:onto:order}.\ref{stm:plan:onto:order:a} and~\ref{stm:plan:onto:order}.\ref{stm:plan:onto:order:b}.

Given a conflict-free path-based plan $\{\pi_a\}_{a\in A}$ for a MAPF problem $(V,E,A)$,
the corresponding minimal conflict-free path-based ordered event set
is the smallest ordered event set
\(
(\{a@\pi_a(i)\mid a\in A, 0\leq i\leq|\pi_a|\}, {\prec})
\)
satisfying the following conditions:
\begin{enumerate}
\item ${a}@{\pi_a(i)}\prec{a}@{\pi_a(i+1)}$ for all $a\in A$ and $0 \leq i < n$ with $\pi_a(i) \neq \pi_a(i+1)$, and
\item ${a}@{\pi_a(i+1)}\prec{b}@{\pi_b(j)}$
  for all $a, b \in A$ and $0 \leq i < j \leq n$ with $a \neq b$ and $\pi_a(i) = \pi_b(j) \neq \pi_a(i+1)$.
\end{enumerate}

Observe that the ordered event set in~\eqref{ex:event:set:conflict:free}
satisfies the above conditions for plan $\{\pi_a^6,\pi_b^6\}$ given in~\eqref{eq:path:b:six}.
In fact, it is the smallest ordered event set satisfying them.

Given a conflict-free path-based ordered event set $(\events, {\prec})$ for a MAPF problem $(V,E,A)$,
a conflict-free path-based plan $\{\pi_a\}_{a\in A}$ is constructed as follows.
First of all, we define a sequence of mappings from events to time steps:
for each $\epsilon \in \events$ and $i > 0$, define
\begin{enumerate}
\item $\alpha_0(\epsilon) = 0$
\item
  \(
  \alpha_i(\epsilon) =
  \max \{ \alpha_{i-1}(\epsilon)                                                                    \} \cup
       \{ \alpha_{i-1}(\epsilon')+1 \mid \epsilon' \in \events \text{, }\epsilon' \precdot \epsilon \}
  \)
\end{enumerate}
Consider a fixed point $\alpha = \alpha_i$ of this construction satisfying $\alpha_i=\alpha_{i+1}$ for some $i\geq 0$.
Intuitively, $\alpha$ gives the earliest arrival time of each agent at its traversed locations.
With it, we define the stroll $\pi_a=(u_i)_{i=0}^n$ of each agent $a\in A$ as follows:
\begin{enumerate}
\item $n = \max\{ \alpha(\epsilon) \mid \epsilon \in \events \}$
\item $u_j = v$ for all $a@v \precdot a@w $ and all $\alpha(a@v) \leq j < \alpha(a@w)$
\item $u_j = v$ for $a@v = \max_{\prec_a}{\events_a}$ and all $\alpha(a@v) \leq j \leq n$
\end{enumerate}
The resulting plan $\{\pi_a\}_{a\in A}$ is conflict-free and path-based.
Note that there is no shorter plan corresponding to the order since agents move as early as possible.

\begin{table}
\caption{Arrival time mapping construction for the ordered event set in~\eqref{ex:event:set:conflict:free}}\label{tab:arrive}
\centering
\begin{tabular}{rrrrrrrrrr}
  \hline\hline
  \multicolumn{1}{c}{$\alpha_i$} & \multicolumn{5}{c}{agent $a$} & \multicolumn{4}{c}{agent $b$}
  \\\cline{1-1}\cline{2-6}\cline{7-10}
  \multicolumn{1}{c}{$i$} & \pos{0}{2} & \pos{0}{1} & \pos{1}{1} & \pos{1}{2} & \pos{1}{3} & \pos{1}{0} & \pos{1}{1} & \pos{0}{1} & \pos{0}{0} \\
  \hline
  0 & 0 & 0 & 0 & 0 & 0 & 0 & 0 & 0 & 0 \\
  1 & 0 & 1 & 1 & 1 & 1 & 0 & 1 & 1 & 1 \\
  2 & 0 & 1 & 2 & 2 & 2 & 0 & 2 & 2 & 2 \\
  3 & 0 & 1 & 2 & 3 & 3 & 0 & 3 & 3 & 3 \\
  4 & 0 & 1 & 2 & 3 & 4 & 0 & 4 & 4 & 4 \\
  5 & 0 & 1 & 2 & 3 & 4 & 0 & 4 & 5 & 5 \\
  6 & 0 & 1 & 2 & 3 & 4 & 0 & 4 & 5 & 6 \\
  \hline\hline
\end{tabular}
\end{table}

Consider the construction of arrival time mappings in Table~\ref{tab:arrive} for the ordered event set in~\eqref{ex:event:set:conflict:free}.
Each row corresponds to a mapping,
where the first column is the running index $i$
and the remaining ones give the arrival time of an agent at a vertex as indicated by the table header.
We obtain that $\alpha = \alpha_6$ is a fixed point.
Observe that we can use this mapping to construct the plan $\{\pi_a^6,\pi_b^6\}$ given in~\eqref{eq:path:b:six}.
To verify, note that $\min \iota_c(v) = \alpha(v@c)$ for all $c \in \{ a, b \}$ and $v \in \pi_c^6$.

Arrival time mappings like $\alpha$ play a major role when dealing with durations in Section~\ref{sec:scheduling}.

 \subsection{Fact format}\label{sec:mapf:format}

We represent a MAPF problem $(V,E,A)$ as a set of facts $\Gamma(V,E,A)$
consisting of
an atom \ASP|vertex($v$)| for each vertex $v\in V$,
an atom \ASP|edge($u$,$v$)| for each edge   $(u,v)\in E$, and
atoms \ASP|agent($a$)|, \ASP|start($a$,$s_a$)|, and \ASP|goal($a$,$g_a$)| for each agent $a\in A$.
We assume that a suitable syntactic representation is chosen for values of italic variables;
this representation is set in typewriter in the source code.

We represent plans by predicates
\ASP|move/3| or \ASP|move/4|,
depending on whether we use explicit time points or not.
In both cases,
the first argument identifies an agent,
and the second and third an edge.
A path or stroll $(v_i)_{i=0}^n$ of an agent $a$ is then represented by atoms of form
\ASP|move($a$,$v_{i-1}$,$v_i$)| or \ASP|move($a$,$v_{i-1}$,$v_i$,$i$)|
with $v_{i-1} \neq v_i$ for $0 < i \leq n$, respectively.
We use the variant with time points in the next section
and drop them in Section~\ref{sec:mapf:encoding:ordering}.

\subsection{Vanilla Encoding for MAPF}
\label{sec:mapf:encoding:vanilla}

We first present a basic encoding for MAPF following the traditional approach in Answer Set Planning~\cite{lifschitz02a},
which relies on time steps.

The encoding in Listing~\ref{lst:mapf:vanilla} computes conflict-free plans~$\{\pi_a\}_{a\in A}$ of length $n$ for MAPF problems~$(V,E,A)$;
its parameter~$m$ allows for preventing follow conflicts when set to \ASP|fc|.
\begin{lstlisting}[language=clingoht,
                   caption={Encoding to find bounded length plans for MAPF},
                   label={lst:mapf:vanilla}]
{ move(A,U,V,T): edge(U,V) } <= 1 :- agent(A), T=1..$n$.|\label{lst:mapf:vanilla:move}|

at(A,U,0) :- start(A,U).|\label{lst:mapf:vanilla:init}|
at(A,V,T) :- move(A,_,V,T), T=1..$n$.|\label{lst:mapf:vanilla:effect}|
at(A,U,T) :- at(A,U,T-1), not move(A,U,_,T), T=1..$n$.|\label{lst:mapf:vanilla:inertia}|

:- move(A,U,_,T), not at(A,U,T-1).|\label{lst:mapf:vanilla:valid}|
:- goal(A,U), not at(A,U,$n$).|\label{lst:mapf:vanilla:goal}|

:- { at(A,U,T) } > 1, vertex(U), T=0..$n$.|\label{lst:mapf:vanilla:vconflict}|
:- move(_,U,V,T), move(_,V,U,T).|\label{lst:mapf:vanilla:sconflict}|
:- at(A,U,T), at(B,U,T+1), A!=B, $m$=fc.|\label{lst:mapf:vanilla:fconflict}|

:- { at(A,U,T) } != 1, agent(A), T=1..$n$.|\label{lst:mapf:vanilla:loc:unique}|
\end{lstlisting}

We use a choice rule in Line~\ref{lst:mapf:vanilla:move} to generate move candidates.
For each time point $1 \leq t \leq n$ and each agent $a \in A$,
we choose at most one atom \ASP|move($a$,$u$,$v$,$t$)| for an edge $(u,v) \in E$.
The selected moves indicate agents $a$ exiting vertex~$u$ at time point~$t-1$ and arriving at vertex~$v$ at time point~$t$.

In Lines~\ref{lst:mapf:vanilla:init}, \ref{lst:mapf:vanilla:effect}, and~\ref{lst:mapf:vanilla:inertia},
we generate candidates for strolls $\pi_a$ from the start positions~$s_a$ of agents $a \in A$ and moves selected in Line~\ref{lst:mapf:vanilla:move}.
Line~\ref{lst:mapf:vanilla:init} ensures that $\pi_a(0) = s_a$ encoded by \ASP|at($a$,$s_0$,0)|.
In the following line,
we derive agent positions \ASP|at($a$,$v$,$t$)| from \ASP|move($a$,$u$,$v$,$t$)| establishing $\pi_a(t) = v$.
The last line in the block encodes that an agent stays at its position if it has not been moved.
That is, if we have $\pi_a(t-1) = u$ and the agent is not moved at time point~$t$,
we obtain $\pi_a(t) = u$ captured by \ASP|at($a$,$u$,$t$)|.
Note the use of \ASP|_| in the negative literal.
This can be seen as a shortcut for \ASP|not move(A,U,T)| together with the rule \ASP|move(A,U,T) :- move(A,U,V,T)| projecting out variable~\ASP|V|.

At this point,
the vertices in the stroll candidates~$\pi_a$ are not necessarily connected by edges nor do they lead to goal vertices~$g_a$.
This is taken care of by the integrity constraints in Lines~\ref{lst:mapf:vanilla:valid} and~\ref{lst:mapf:vanilla:goal}.
The former ensures connectedness.
We discard candidates whenever the agent is not at the source vertex of a move.
For each true atom \ASP|move($a$,$u$,$v$,$t$)|,
we require $\pi_a(t-1) = u$ by discarding candidate solutions not including \ASP|at($a$,$u$,$t-1$)|.
Now, stroll candidates are indeed strolls.
The only missing piece is to require that they lead to goal vertices.
This is addressed by the second constraint ensuring that $\pi_a(n) = g_a$.

For now,
we have a plan $\{\pi_a\}_{a\in A}$ that is not necessarily conflict-free.
In Lines~\ref{lst:mapf:vanilla:vconflict}, \ref{lst:mapf:vanilla:sconflict}, and~\ref{lst:mapf:vanilla:fconflict},
we ensure that the plan is free of vertex, swap, and follow conflicts, respectively.
The first integrity constraint discards plans in which two agents occupy the same vertex at the same time.
The next one ensures that there are no swap conflicts.
Here, we slightly deviate from the definition.
A swap conflict occurs if two agents travel between two vertices in opposing directions at the same time point.
Unlike this, we discard solutions with opposing moves.
We treat follow conflicts according to the definition in Line~\ref{lst:mapf:vanilla:fconflict}.

At last,
we add a redundant check in Line~\ref{lst:mapf:vanilla:loc:unique}
asserting that any agent must be at exactly one position.
This check is meant to improve solving performance.

Finally,
we establish the following correspondence between the stable models of our vanilla encoding along with facts representing a MAPF problem
and conflict-free plans for the same MAPF problem.
\begin{proposition}[Soundness]\label{prop:vanilla:sound}
  Given a MAPF problem $(V,E,A)$,
  let $P$ be the union of\/ $\Gamma(V,E,A)$ and the encoding in Listing~\ref{lst:mapf:vanilla}
  for some $n\in\mathbb{N}_0$ (and $m=\text{\ASP|fc|}$).

  If $X$ is a stable model of $P$, then
  \(
  \{(v_i \mid \text{\ASP|at($a$,$v_i$,$i$)|}\in X)_{i=0}^n\}_{a\in A}
  \)
  is a vertex, swap, (and follow) conflict-free plan for $(V,E,A)$.
\end{proposition}
Thus,
each sequence
\(
(v_i\mid \text{\ASP|at($a$,$v_i$,$i$)|}\in X)_{i=0}^n
\)
of vertices
is a stroll from $s_a$ to $g_a$ in $(V,E)$ for all $a\in A$.
\begin{proposition}[Completeness]\label{prop:vanilla:complete}
  Given a MAPF problem $(V,E,A)$,
  let $P$ be the union of\/ $\Gamma(V,E,A)$ and the encoding in Listing~\ref{lst:mapf:vanilla}
  for some $n\in\mathbb{N}_0$ (and $m=\text{\ASP{fc}}$).

  If \(\{\pi_a\}_{a\in A}\) is a vertex, swap (and follow) conflict-free plan for $(V,E,A)$,
  then
  \(
  \Gamma(V,E,A) \cup
  \{ \text{\ASP|at($a$,$\pi_a(i)$,$i$)|} \mid a\in A \text{,\ } 0\leq i \leq n \} \cup
  \{ \text{\ASP|move($a$,$\pi_a(i-1)$,$\pi_a(i)$,$i$)|} \mid 0 < i \leq n\text{,\ } \pi_a(i-1) \neq \pi_a(i) \}  \)
  is a stable model of $P$.
\end{proposition}

\subsection{Ordering Encoding for MAPF}
\label{sec:mapf:encoding:ordering}

We now present a solution to MAPF reflecting event orderings;
it consists of Listing~\ref{lst:mapf:path}, \ref{lst:mapf:order}, and~\ref{lst:mapf:check}.
The underlying encoding technique
relies on acyclicity constraints and
drops time steps and explicit bounds on the length of plans.

\begin{lstlisting}[language=clingoht,
                   caption={Encoding to find candidate paths for MAPF problems},
                   label={lst:mapf:path}]
{ move(A,U,V): edge(U,V) } <= 1 :- agent(A), vertex(V).|\label{lst:mapf:path:upper:in}|
{ move(A,U,V): edge(U,V) } <= 1 :- agent(A), vertex(U).|\label{lst:mapf:path:upper:out}|
:-  move(A,U,_), not start(A,U), not move(A,_,U).|\label{lst:mapf:path:lower:in}|
:-  move(A,_,U), not  goal(A,U), not move(A,U,_).|\label{lst:mapf:path:lower:out}|

:- start(A,U), move(A,_,U).|\label{lst:mapf:path:start:in}|
:-  goal(A,U), move(A,U,_).|\label{lst:mapf:path:goal:out}|
:- start(A,U), not  goal(A,U), not move(A,U,_).|\label{lst:mapf:path:goal:in}|
:-  goal(A,U), not start(A,U), not move(A,_,U).|\label{lst:mapf:path:start:out}|
\end{lstlisting}
The first encoding in Listing~\ref{lst:mapf:path} selects atoms \ASP|move($a$,$u$,$v$)|
for agents~$a \in A$ and edges~$(u,v) \in E$ for a MAPF problem~$(V,E,A)$.
Given a stable model $X$ of Listing~\ref{lst:mapf:path},
the selected moves for each agent $a$ form a subgraph~$(V_a,E_a)$ of~$(V,E)$ such that\footnote{Functions $\indegree(v)$ and $\outdegree(v)$ give the in and out degree of vertex $v\in V_a$ in $(V_a,E_a)$.}
\begin{enumerate}
\item $E_a = \{ (u,v) \mid \text{\ASP|move($a$,$u$,$v$)|} \in X \}$,
\item $V_a = \{s_a, g_a\} \cup \{ u, v \mid (u,v) \in E_a\}$,
\item $\indegree(v) = \outdegree(u) = 1$ for all $\text{\ASP|move($a$,$u$,$v$)|}\in X$,
\item $\indegree(u) = 1$ if $u \neq s_a$ and $\outdegree(v) = 1$ if $v \neq g_a$ for all $\text{\ASP|move($a$,$u$,$v$)|}\in X$,
\item $\indegree(s_a) = \outdegree(g_a) = 0$, and
\item $\indegree(g_a) = \outdegree(s_a) = 1$ if $s_a \neq g_a$.
\end{enumerate}
Any vertex on a path different from the start vertex must have an in  degree ($\indegree$)  of one in $(V_a,E_a)$;
any vertex on a path different from the goal  vertex must have an out degree ($\outdegree$) of one in $(V_a,E_a)$;
the start and goal vertices must be the start and end of a path.
Hence, the subgraph for an agent~$a$ consists of exactly one path leading from $s_a$ to $g_a$ and
zero or more separate cycles with a length greater than or equal to two.

To see this,
let us go over the rules and see how they affect the in and out degrees of vertices.
The rules in Lines~\ref{lst:mapf:path:upper:in} and~\ref{lst:mapf:path:upper:out} generate atoms \ASP|move($a$,$u$,$v$)| for agents $a \in A$ such that $(u,v) \in E$.
Furthermore, they ensure that $\indegree(v) \leq 1$ and $\outdegree(u) \leq 1$ for all such \ASP|move($a$,$u$,$v$)|.
We also have $\indegree(v) \geq 1$ and $\outdegree(u) \geq 1$ for all \ASP|move($a$,$u$,$v$)|.
Thus, at this point, we have $\indegree(v) = \outdegree(u) = 1$ whenever \ASP|move($a$,$u$,$v$)| is generated.
Lines~\ref{lst:mapf:path:lower:in} and~\ref{lst:mapf:path:lower:out} ensure that
$\indegree(u) = 1$ if $u \neq s_a$ and $\outdegree(v) = 1$ if $v \neq g_a$ for all \ASP|move($a$,$u$,$v$)|.
Note that this uses $\indegree(u) \leq 1$ and $\outdegree(v) \leq 1$ enforced in Lines~\ref{lst:mapf:path:upper:in} and~\ref{lst:mapf:path:upper:out}.
Lines~\ref{lst:mapf:path:start:in} and~\ref{lst:mapf:path:goal:out} make sure that $\indegree(s_a) = \outdegree(g_a) = 0$.
Lines~\ref{lst:mapf:path:goal:in} and~\ref{lst:mapf:path:start:out} establish that $\indegree(g_a) = \outdegree(s_a) = 1$ if $s_a \neq g_a$.
This also uses $\indegree(g_a) \leq 1$ and $\outdegree(s_a) \leq 1$ warranted by Lines~\ref{lst:mapf:path:upper:in} and~\ref{lst:mapf:path:upper:out}.

The second encoding in Listing~\ref{lst:mapf:order} selects atoms \ASP|resolve($a$,$b$,$u$)| in accordance with Definition~\ref{def:order:conflict-free}.
When such an atom is derived, agent $a$ has to depart from vertex $u$ before agent $b$ arrives at vertex $u$.
\begin{lstlisting}[language=clingoht,
                   caption={Encoding to find candidate event orders for MAPF problems},
                   label={lst:mapf:order}]
resolve(A,B,U) :- start(A,U), move(B,_,U), A!=B.|\label{lst:mapf:order:start}|
resolve(A,B,U) :-  goal(B,U), move(A,_,U), A!=B.|\label{lst:mapf:order:goal}|
{ resolve(A,B,U);
  resolve(B,A,U) } >= 1 :- move(A,_,U), move(B,_,U), A<B.|\label{lst:mapf:order:move}|

:- resolve(A,B,U), resolve(B,A,U).|\label{lst:mapf:order:unique}|
\end{lstlisting}
The rule in Line~$\ref{lst:mapf:order:move}$ chooses which of two agents moving to the same vertex has to move first.
The encoding assumes that $s_a \neq s_b$ and $g_a \neq g_b$ for all $a,b\in A$ with $a \neq b$.
Thus, a conflict at a start and goal vertex can only arise if another agent moves to such a vertex.
The rules in Lines~$\ref{lst:mapf:order:start}$ and~$\ref{lst:mapf:order:goal}$ select the right resolution order at these vertices:
agents at their start vertex as well as
agents passing through the goal position of another agent have to move first.
At this point, there is at least one \ASP|resolve| atom for each case in Definition~\ref{def:order:conflict-free}.
The constraint in Line~\ref{lst:mapf:order:unique} ensures that it is only one.

Note that there is exactly one atom \ASP|move($a$,$u$,$v$)| for each obtained \ASP|resolve($a$,$b$,$u$)|.
This is enforced for atoms derived by the rule in Line~\ref{lst:mapf:order:move}
by the integrity constraint in Line~\ref{lst:mapf:path:lower:out} in Listing~\ref{lst:mapf:path}.
Assume that \ASP|resolve($a$,$b$,$u$)| is derived by the rule in Line~\ref{lst:mapf:order:start}.
Then there are two cases.
If $s_a =    g_a$, a conflicting atom would be derived in Line~\ref{lst:mapf:order:goal}.
If $s_a \neq g_a$, there must be a move for agent~$a$.
The same argument can be made for \ASP|resolve| atoms derived by the rule in Line~\ref{lst:mapf:order:goal}.

Finally, the third and last encoding in Listing~\ref{lst:mapf:check} ensures that the \ASP|move| and \ASP|resolve| atoms
from Listings~\ref{lst:mapf:path} and~\ref{lst:mapf:order}, respectively, form a conflict-free path-based ordered event set.\footnote{We consider alternative encodings for this in plain ASP as well as via difference constraints in Section~\ref{sec:experiments}.}
\begin{lstlisting}[language=clingoht,
                   caption={Encoding to find conflict-free ordered event sets for MAPF},
                   label={lst:mapf:check}]
#edge ((A,U),(A,V)) : move(A,U,V).|\label{lst:mapf:check:path}|
#edge ((A,V),(B,U)) : resolve(A,B,U), move(A,U,V).|\label{lst:mapf:check:conflict}|
\end{lstlisting}
The edge directive in Line~\ref{lst:mapf:check:path} specifies a graph containing edges given by the \ASP|move| atoms.
We can interpret the tuples $(a,u)$ in the edge right after the \ASP|#edge| keyword as events $a@u$.
Thus for each true~\ASP|move($a$,$u$,$v$)| an edge~$(a@u,a@v)$ is added to the graph.
The solver discards all solutions where this graph contains a cycle.
Remember that the encoding in Listing~\ref{lst:mapf:path} admits graphs with paths and cycles for agents.
Thus, at this point, we have ensured that the moves form a path-based ordered event set.
The edge directive in Line~\ref{lst:mapf:check:conflict} further extends the above graph.
In accord with Definition~\ref{def:order:conflict-free},
we add edge $(a@v,b@u)$ for each atom \ASP|resolve($a$,$b$,$u$)|
where $a@u \precdot_a a@v$ is captured by \ASP|move($a$,$u$,$v$)|.
We have argued above that there is exactly one such move.
Thus,
if the resulting graph is acyclic,
then there is a corresponding conflict-free path-based ordered event set.

In the remainder of this section,
we show how to obtain event orders from stable models of the above logic programs and vice versa.
\begin{definition}\label{def:mapf:model-to-order}
  Let $(V,E,A)$ be a MAPF problem and $X$ be a stable model of the union of\/ $\Gamma(V,E,A)$ and
  the encodings in Listings~\ref{lst:mapf:path}, \ref{lst:mapf:order}, and~\ref{lst:mapf:check}.

  The ordered event set $(\events, \prec)$ corresponding to $X$
  is the smallest ordered event set such that
  \begin{enumerate}
  \item $\events = \{ a@u, a@v \mid \text{\ASP|move($a$,$u$,$v$)|} \in X \} \cup \{ a@u \mid \text{\ASP|start($a$,$u$)|} \in X \}$
  \item $a@u \prec a@v$ for all $\text{\ASP|move($a$,$u$,$v$)|} \in X$, and
  \item $a@v \prec b@u$ for all $\text{\ASP|resolve($a$,$b$,$u$)|}, \text{\ASP|move($a$,$u$,$v$)|} \in X$.
  \end{enumerate}
\end{definition}
The second part of \events\ addresses non-moving agents having identical start and goal positions.
\begin{proposition}[Soundness]\label{stm:mapf:model-to-order}
  Let $(V,E,A)$ be a MAPF problem and $P$ be the union of\/ $\Gamma(V,E,A)$ and
  the encodings in Listings~\ref{lst:mapf:path}, \ref{lst:mapf:order}, and~\ref{lst:mapf:check}.

  If $X$ is a stable model of $P$,
  then the ordered event set $(\events, \prec)$ corresponding to $X$ is
  a minimal conflict-free path-based ordered event set for $(V,E,A)$.
\end{proposition}
Together with Proposition~\ref{stm:plan:onto:order}.\ref{stm:plan:onto:order:b},
this implies the existence of a corresponding path-based plan,
which can be constructed as described at the end of Section~\ref{sec:ordering}.
\begin{definition}\label{def:mapf:order-to-model}
  Let $(V,E,A)$ be a MAPF problem and $(\events,\prec)$ be a minimal conflict-free path-based ordered event set for $(V,E,A)$.

  We define the set $X$ of atoms corresponding to $(\events,\prec)$ as
  the smallest set such that
  \begin{enumerate}
  \item $\Gamma(V,E,A)\subseteq X$,
  \item $\text{\ASP|move($a$,$u$,$v$)|} \in X$ for all $a@u \precdot_a a@v$, and
  \item $\text{\ASP|resolve($a$,$b$,$u$)|} \in X$ for all $a@v \prec b@u$ and $a@u \precdot_a a@v$ with $a \neq b$.
  \end{enumerate}
\end{definition}
\begin{proposition}[Completeness]\label{stm:mapf:order-to-model}
  Let $(V,E,A)$ be a MAPF problem and $P$ be the union of\/ $\Gamma(V,E,A)$ and
  the encodings in Listings~\ref{lst:mapf:path}, \ref{lst:mapf:order}, and~\ref{lst:mapf:check}.

  If $(\events,\prec)$ is a minimal conflict-free path-based ordered event set for $(V,E,A)$,
  then the set of atoms corresponding to $(\events,\prec)$ is a stable model of $P$.
\end{proposition}
This and Proposition~\ref{stm:plan:onto:order}.\ref{stm:plan:onto:order:a} implies that
each path-based plan of a MAPF problem
has a corresponding stable model of the MAPF encodings.

\section{Routing and scheduling}
\label{sec:scheduling}

We now shift our attention to the combination of routing and scheduling.
To this end,
we begin by extending our key concepts with means for accommodating durations.

A weighted graph is a triple $(V,E,\delta)$,
where $(V,E)$ is a (finite, simple directed) graph
and $\delta:E\rightarrow\mathbb{N}$
maps edges to positive integers.
The function $\delta$ gives the travel duration along each edge.

As in Section~\ref{sec:background},
we define strolls over extended graphs.
To accommodate waiting agents,
we extend graphs with loops $(v,v)$ for vertices $v\in V$ as in the unweighted case.
To account for durations,
we introduce additional vertices and edges
to model agents moving along edges while being in between vertices;
intuitively, an agent located at such an auxiliary vertex~$e^{u,v}_i$ moved for $i$ time units along the edge $(u,v)$.
\begin{definition}\label{def:wstroll}
  A \emph{stroll} $(u_i)_{i=0}^n$ in a weighted graph $(V,E,\delta)$ is a walk in the directed graph $(V',E')$
  where
  \begin{enumerate}
    \item $u_0,u_n \in V$,
    \item $V' = V \cup \{ e^{u,v}_i \mid (u,v) \in E\text{, }1 \leq i < \delta(u,v) \}$, and
    \item $E' = \{ (v,v) \mid v \in V \} \cup \{ \eta((u,v),i) \mid (u,v) \in E, 1\leq i\leq\delta(e) \}$
      \ where
      \[
      \eta((u,v),i) =
      \begin{cases}
      (u,v) & \text{if } \delta(u,v) = 1 \text{ and } i = 1,\\
      (u,e^{u,v}_i) & \text{if } \delta(u,v) > 1 \text{ and } i = 1,\\
      (e^{u,v}_{i-1},e^{u,v}_i) & \text{if } \delta(u,v) > 1 \text{ and } 2 \leq i < \delta(u,v), \text{and}\\
      (e^{u,v}_{i-1},v) & \text{if } \delta(u,v) > 1 \text{ and } i = \delta(u,v).
      \end{cases}
      \]
  \end{enumerate}
  A stroll $(u_i)_{i=0}^n$ is \emph{path-like}, if
  \(
  \iota(u_i)=[\min\iota(u_i),\max\iota(u_i)]
  \)
  for $0\leq i\leq n$.
\end{definition}
This reduction to walks in unweighted graphs keeps agents' moves discrete and synchronous.
Also, it collapses to the original definition of strolls in case we uniformly assign a duration of one.
As with basic strolls, an agent may dwell on an original vertex, but once it leaves, it moves continuously to the target vertex.

\begin{figure}[h]
\figrule \includegraphics{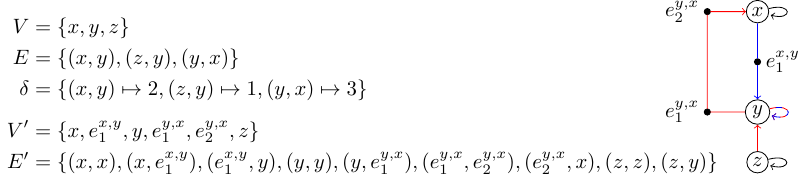}
\caption{A weighted graph and its corresponding unweighted graph}\label{fig:duration:wgraph}
\figrule \end{figure}

As an example,
consider the weighted graph~$(V,E,\delta)$ together with its corresponding unweighted graph $(V',E')$ in Figure~\ref{fig:duration:wgraph}.
Strolls in this weighted graph are defined in terms of walks in the unweighted graph $(V',E')$.
The two possible strolls (of the same length) below are highlighted in blue and red in the figure, respectively:
\begin{align}\label{ex:duration:strolls}
  \pi_1&= (x,e^{x,y}_1,y,y,y,y)&\pi_2&= (z,y,y,e^{y,x}_1,e^{y,x}_2,x)
\end{align}

The next definition captures moves in terms of arrival and departure times at the original vertices of a stroll.
\begin{definition}\label{def:wstroll:move}
  We define the set~$\mathcal{M}_\pi$ of \emph{moves}
  for a stroll~$\pi$ in a weighted graph~$(V,E,\delta)$
  as the set of all pairs~$(i,j)$ with $0 \leq i < j \leq |\pi|$ such that
  $(\pi(i), \pi(j)) \in E$ and $\pi(k) \notin V$ for all $i < k < j$.
\end{definition}

For the two strolls in \eqref{ex:duration:strolls},
we obtain the sets $\mathcal{M}_{\pi_1}=\{(0,2)\}$ and $\mathcal{M}_{\pi_2}=\{(0,1),(2,5)\}$.

A weighted MAPF problem is a quintuple~$(V,E,A,\delta,\sigma)$,
where
$(V,E,A)$      is a MAPF problem,
$(V,E,\delta)$ is a weighted graph, and
$\sigma: E\to\mathbb{N}_0$
gives a safety period among successive visits of a vertex by two agents.
We express this relative to the departing agent, since it generalizes the idea of follow conflicts (see below).
A \emph{plan} of length~$n$ for a weighted MAPF problem~$(V,E,A,\delta,\sigma)$ is a family $\{\pi_a\}_{a \in A}$
of strolls of length~$n$ in $(V,E,\delta)$
such that $\pi_a(0) = s_a$ and $\pi_a(n) = g_a$ for all $a \in A$.
As in Section~\ref{sec:background},
a plan for a weighted MAPF problem is path-based, if all its strolls are path-like.

We lift the three types of conflicts to the weighted case by using the concept of moves.
\begin{definition}\label{def:wplan:conflict}
  A plan $\{\pi_a\}_{a \in A}$ of length $n$ for a weighted MAPF problem~$(V,E,A,\delta,\sigma)$ is
  \begin{enumerate}
  \item \label{def:wplan:vconflict}
    \emph{vertex conflict-free} if
    $\pi_a(i) \neq \pi_b(i)$ for all $a,b \in A$ and $0 \leq i \leq n$ such that $a \neq b$ and $\pi_a(i),\pi_b(i) \in V$,
  \item \label{def:wplan:sconflict}
    \emph{swap conflict-free} if
    $j \notin (k,l]$
    for all $a,b \in A$, $(i,j) \in \mathcal{M}_{\pi_a}$, $(k,l) \in \mathcal{M}_{\pi_b}$ such that $a \neq b$, $\pi_a(i) = \pi_b(l)$, and $\pi_a(j) = \pi_b(k)$,
    and
  \item \label{def:wplan:fconflict}
    \emph{$\sigma$-follow conflict-free} if
    $j \notin (k,k+\sigma(\pi_b(k),\pi_b(l))]$
    for all $a,b \in A$, $(i,j) \in \mathcal{M}_{\pi_a}$, $(k,l) \in \mathcal{M}_{\pi_b}$ such that $a \neq b$ and $\pi_a(j) = \pi_b(k)$.
  \end{enumerate}
\end{definition}
We refer to plans for weighted MAPF problems $(V,E,A,\delta,\sigma)$ as conflict-free
if they are vertex, swap, and $\sigma$-follow conflict-free.

\begin{figure}
\figrule \includegraphics{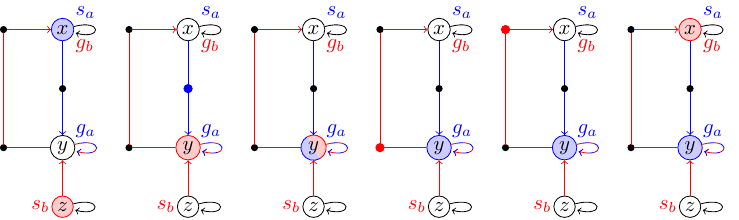}
\caption{Two strolls having a vertex conflict at index 2}\label{fig:duration:strolls}
\figrule \end{figure}

Vertex conflicts remain unchanged from unweighted MAPF,
just that their occurrence on the original vertices must be enforced in the weighted case.
For instance,
let us use the graph from Figure~\ref{fig:duration:wgraph} and introduce two agents $A = \{a, b\}$ with $s_a = x$, $s_b = z$, and $g_a = y$, $g_b = x$.
Taking an arbitrary $\sigma$, we obtain a weighted MAPF problem $(V,E,A,\delta,\sigma)$.
The plan $\{\pi_a, \pi_b\}$ with $\pi_a = \pi_1$ and $\pi_b = \pi_2$ from \eqref{ex:duration:strolls} is a path-based plan for $(V,E,A,\delta,\sigma)$.
As depicted in Figure~\ref{fig:duration:strolls}, we encounter a vertex conflict because $\pi_{a}(2)=\pi_{b}(2) = y$.
As in Section~\ref{sec:ordering},
agent positions are indicated by solid blue and red discs;
larger transparent discs are used for original vertices.

\begin{figure}
\figrule \includegraphics{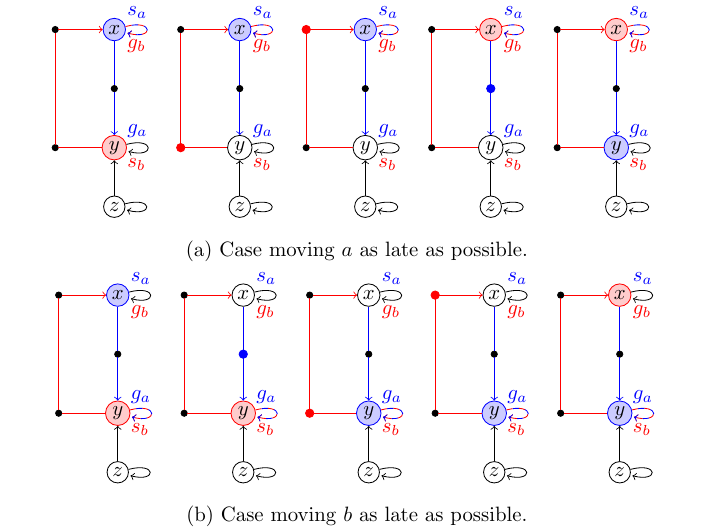}
\caption{Two strolls with swap conflicts, no matter the delay}\label{fig:duration:sconflict}
\figrule \end{figure}

As before,
a swap conflict occurs if two agents move between two vertices in opposite directions at the same time,
just that now both moves may stretch over distinct durations.
Two such moves $(i,j)$ and $(k,l)$ overlap, if either $j \in (k,l]$ or $l \in (i,j]$.
In words,
the arrival of one agent at its destination must not occur while the other agent is on its way to its end point,
traversing intermediate nodes leading to it.
For instance,
let us consider our weighted MAPF example
with adjusted start and goal positions for agents $a$ and $b$
as depicted in Figure~\ref{fig:duration:sconflict}.
First of all,
we note that the two agents have several ways to cross from one vertex to another without causing a vertex conflict.
However,
no matter whether we delay $a$ as much as possible after moving $b$, viz. Figure~\ref{fig:duration:sconflict}a),
or delay $b$ as much as possible after moving $a$, viz. Figure~\ref{fig:duration:sconflict}b),
we obtain swap conflicts.
In the first case,
we have move~$(2,4)$ for agent~$a$ and~$(0,3)$ for~$b$
resulting in a swap conflict because $3\in (2,4]$.
Similarly, in the second case,
we have move~$(0,2)$ for agent~$a$ and~$(1,4)$ for~$b$,
also yielding a swap conflict because $2\in (1,4]$.

\begin{figure}[ht]
\figrule \includegraphics{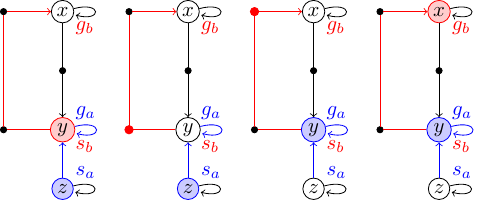}
\caption{Two strolls having a follow conflict with safety period $\sigma(y,x)=2$, but none with $\sigma(y,x)=1$}\label{fig:duration:fconflict}
\figrule \end{figure}

The idea of $\sigma$-follow conflicts is to put safety periods between the departure of one agent and the
arrival of the next agent at the same vertex.
As an example,
consider the moves $(1,2)$ and $(0,3)$ of agents $a$ and $b$ in Figure~\ref{fig:duration:fconflict}.
For $\sigma(y,x)=2$, we get a follow conflict because $2\in (0,2]$.
However, once the security distance is reduced by one, viz.\ $\sigma(y,x)=1$,
the plan is follow conflict-free because $2 \notin (0,1]$.
Observe that setting $\sigma$ to zero allows an agent to enter a vertex at the same time as another agent leaves that vertex.
Setting $\sigma$ to one forces an agent to wait one time point before entering a vertex just left by another agent.
Furthermore,
if the value of $\sigma$ is everywhere greater or equal to that of $\delta$,
the absence of $\sigma$-follow conflict implies swap conflict-freeness.
However, conversely,
follow conflict-freeness for other choices of $\sigma$ may not imply swap conflict-freeness.
All in all, the addition of safety periods results in a rather versatile concept of follow conflicts;
at the same, it must be handled with care since its relationship to other types of conflicts does not carry
over from the unweighted case in general.

Lastly, if all edge weights are one,
we can relate weighted and unweighted MAPF problems:
Depending on whether $\sigma$ is zero or one everywhere,
plans for weighted MAPF problems correspond to plans for unweighted ones with or without follow conflicts, respectively,
provided that they are vertex conflict-free.\footnote{Vertex conflict-freeness is a prerequisite since follow conflict-freeness is weaker in the weighted case.}

In the following, we consider three refined variants of $\sigma$-follow conflicts:
\begin{enumerate}
\item \emph{vertex-follow} conflicts prevent an agent to follow another one
  until it reaches its target vertex one time point later:
  $\sigma_{\mathrm{v}}(u,v) = \delta(u,v)$.

\item \emph{edge-follow} conflicts prevent an agent to follow another one
  until it reaches its target vertex:
  $\sigma_{\mathrm{e}}(u,v) = \delta(u,v)-1$.

\item \emph{$d$-safety-follow} conflicts for non-negative integers $d$ prevent agents to follow one another
  until $d$ time units have elapsed:
  $\sigma_{\mathrm{s},d}(u,v) = d$.
\end{enumerate}
We can once more compare to unweighted MAPF by assuming that
all edge weights are one:
vertex-follow conflicts prevent follow- (and swap-)conflicts in unweighted MAPF.
Edge-follow conflicts prevent two agents from traversing an edge at the same time;
they tolerate follow-conflicts in unweighted MAPF.
Finally, $\sigma_{\mathrm{s},0}$ permits follow conflicts and $\sigma_{\mathrm{s},1}$ prevents follow conflicts.
Also, note that $\sigma_{\mathrm{s},0}$ essentially permits follow-conflicts no matter the edge weights.

 \subsection{Event mappings}\label{sec:mapping}

In what follows,
we are interested in characterizing path-based plans of weighted MAPF problems
in terms of arrival times of agents at vertices.
This generalizes the concept of event orderings from Section~\ref{sec:ordering} to event mappings,
expressed in terms of arrival time mappings.
The definitions of events and event sets remain unchanged, no matter whether the underlying MAPF problem is weighted or not.

We capture arrival times of agents at vertices by mapping events to non-negative integers.
\begin{definition}\label{def:amap}
  Given the event set~$\events$ for a weighted MAPF problem,
  we define~$\alpha:\events\to\mathbb{N}_0$ as an \emph{arrival time mapping} for $\events$.
\end{definition}
Given such an arrival time mapping $\alpha$,
we define the relation~${\prec_\alpha}\subseteq {\events\times\events}$ such that
$\epsilon \prec_{\alpha} \epsilon'$ if $\alpha(\epsilon) < \alpha(\epsilon')$ for $\epsilon,\epsilon' \in \events$.
Note that $\prec_\alpha$ is a total preorder.
Furthermore, we let $\precdot_\alpha$ denote the \emph{cover} of $\prec_\alpha$
noting that $\prec_\alpha$ is transitive by construction.

The \emph{restriction} of an arrival time mapping~$\alpha$ for the event set~$\events$ for a weighted MAPF problem~$(V,E,A,\delta,\sigma)$ to a single agent~$a\in A$
is given by
$\alpha_a:\events_a\to\mathbb{N}_0$ such that
$\alpha_a(\epsilon)=\alpha(\epsilon)$ for all $\epsilon \in \events_a$.

Next, we define arrival time mappings comprising path-like structures.
\begin{definition}\label{def:amap:path-based}
  An arrival time mapping~$\alpha$ for the event set~$\events$ for a weighted MAPF problem~$(V,E,A,\delta,\sigma)$ is \emph{path-based}
  if
  \begin{enumerate}
  \item $\prec_{\alpha_a}$ is a total order with least and greatest elements~$a@s_a$ and~$a@g_a$ for all~$a \in A$, and
  \item $(u,v) \in E$ and $\alpha(a@u) + \delta(u,v) \leq \alpha(a@v)$ for all $a@u, a@v\in\events$ with $a@u \precdot_{\alpha_a} a@v$.
  \end{enumerate}
\end{definition}
This definition is the weighted counterpart of path-based event sets defined in Definition~\ref{def:order:path-based}.

In fact, in analogy to Section~\ref{sec:ordering},
we can associate each path-based arrival time mapping $\alpha$ for some event set \events\
with a path-based plan $\{\pi_a\}_{a\in A}$ for a weighted MAPF problem $(V,E,A,\delta,\sigma)$.
To this end,
we use the latest arrival time of an agent at a vertex as the length $n=\max\{ \alpha(\epsilon) \mid \epsilon \in \events \}$ of the plan
and derive \emph{departure times} from the arrival time mapping:
\begin{align*}
\beta(a@u) &= \begin{cases}
\alpha(a@v) - \delta(u,v) & \text{if there is an $a@v\in \events$ such that $a@u \precdot_{\alpha_a} a@v$, and}\\
n & \text{otherwise.}
\end{cases}
\end{align*}
With the arrival and departure times at hand,
we establish the individual strolls $\pi_a = (u_i)_{i=0}^n$ of the agents $a \in A$ at time points $0 \leq i \leq n$
such that
\begin{align*}u_i &= \begin{cases}
u & \text{$\alpha(a@u) \leq i \leq \beta(a@u)$ and $a@u\in\events$,}\\
e^{u,v}_{i-\beta(a@u)} & \text{$\beta(a@u) < i < \alpha(a@v)$ if $a@u \precdot_{\alpha_a} a@v $.}
\end{cases}
\end{align*}
Observe that this even results in a one-to-one correspondence
if we restrict our attention to path-based plans without unnecessary waits at goal vertices
(cf.\ Proposition~\ref{stm:wplan:amap}).
\comment{T:\par
  $a@u \precdot_{\alpha_a} a@v$
  \par implies\par
  $(\beta(a@u),\alpha(a@v)\in\mathcal{M}_{\pi_a}$
  \par
  $(i,j)\in\mathcal{M}_{\pi_a}$
  \par implies\par
  $a@{\pi_a(i)} \precdot_{\alpha_a} a@{\pi_a(j)}$
}

This correspondence allows us to specialize Definition~\ref{def:wplan:conflict} for path-based arrival time mappings.
We say that a path-based arrival time mapping~$\alpha$ for the event set $\events$ for a weighted MAPF problem~$(V,E,A,\delta,\sigma)$ is
\begin{enumerate}
\item\label{def:amap:vconflict}
  \emph{vertex conflict-free} if
  $\alpha(a@v) \notin [\alpha(b@v),\beta(b@v)]$ for all $a@v,b@v \in \events$ such that $a\neq b$,
\item\label{def:amap:sconflict}
  \emph{swap conflict-free} if
  $\alpha(a@v) \notin (\beta(b@v),\alpha(b@u)]$
  for all $a@u,a@v,b@v,b@u \in \events$
  such that $a\neq b$,
  $a@u \precdot_{\alpha_a} a@v$ and $b@v \precdot_{\alpha_b} b@u$,
  and
\item\label{def:amap:fconflict}
  \emph{$\sigma$-follow conflict-free} if
  $\alpha(a@v) \notin (\beta(b@v),\beta(b@v)+\sigma(v,w)]$
  for all $a@u,a@v,b@v,b@w \in \events$
  such that $a\neq b$,
  $a@u \precdot_{\alpha_a} a@v$ and $b@v \precdot_{\alpha_b} b@w$.
\end{enumerate}
As with plans,
we refer to arrival time mappings for weighted MAPF problems $(V,E,A,\delta,\sigma)$ as conflict-free
if they are vertex, swap, and $\sigma$-follow conflict-free.

\begin{table}
\caption{Arrival time mappings producing vertex, swap, and follow conflicts}\label{tab:amap:conflicts}
\newcommand\cc[1]{\cellcolor{#1}}
\centering
\begin{tabular}{llllllllp{5mm}lllllll}
\hline\hline
& $\alpha$ & $0$ & $1$ & \cc{$2$} & 3 & 4 & 5   &  & $\alpha$ & $0$ & $1$ & $2$ & \cc{$3$} & $4$ & \\
\cline{2-2}
\cline{3-8}
\cline{10-10}
\cline{11-15}
& $a$      & $x$ &     & \cc{$y$} &   &   &     &  & $a$      & $x$ &     &     & \cc{   } & $y$ & \\
& $b$      & $z$ & $y$ & \cc{   } &   &   & $x$ &  & $b$      & $y$ &     &     & \cc{$x$} &     & \\
\cline{2-8}
\cline{10-15}
\noalign{\vspace{0.5ex}}
&\multicolumn{7}{l}{(a) The vertex conflict in Figure~\ref{fig:duration:strolls}} & &
\multicolumn{6}{l}{(b) The swap conflict in Figure~\ref{fig:duration:sconflict}a)} & \\
\hline
& $\alpha$ & $0$ & \cc{$1$} & $2$ & $3$ &  &  &  & $\alpha$ & $0$ & $1$ & \cc{$2$} & $3$ & $4$ & \\
\cline{2-2}
\cline{3-8}
\cline{10-10}
\cline{11-15}
& $a$      & $z$ & \cc{$y$} &     &     &  &  &  & $a$      & $x$ &     & \cc{$y$} &     &     & \\
& $b$      & $y$ & \cc{   } &     & $x$ &  &  &  & $b$      & $y$ &     & \cc{   } &     & $x$ & \\
\cline{2-8}
\cline{10-15}
\noalign{\vspace{0.5ex}}
&\multicolumn{7}{l}{\hspace{-3pt}(d) The follow conflict in Figure~\ref{fig:duration:fconflict}} & &
\multicolumn{6}{l}{(c) The swap conflict in Figure~\ref{fig:duration:sconflict}b)} \\
\hline\hline
\end{tabular}
\end{table}

For instance, there is a vertex conflict in Figure~\ref{fig:duration:strolls} with
the arrival time mappings of agents $a$ and $b$ in Table~\ref{tab:amap:conflicts}a).
To see this, observe that
\begin{align*}
\alpha(a@y) = 2 & \in \{ 1,2 \} = [1,2] = [\alpha(b@y),\beta(b@y)]\text{.}
\end{align*}

For another example, consider the graphs in Figure~\ref{fig:duration:sconflict} along with the arrival time mappings in
Table~\ref{tab:amap:conflicts}b) and c).
Again,
the arrival times mappings yield swap conflicts in Figure~\ref{fig:duration:sconflict}a)
because
\begin{align*}
\alpha(b@x) = 3 &\in \{3,4\} = (2,4] = (\beta(a@x),\alpha(a@y)]
\intertext{as well as in Figure~\ref{fig:duration:sconflict}b) because}
\alpha(a@y) = 2 &\in \{ 2,4 \} = (1,4] = (\beta(b@y),\alpha(b@x)]\text{.}
\end{align*}

Finally, given the arrival time mapping in Table~\ref{tab:amap:conflicts}d),
there is a $\sigma$-follow conflict for $\sigma(y,z)=2$ in Figure~\ref{fig:duration:fconflict} because
\begin{align*}
\alpha(a@y) = 1 & \in \{ 1,2 \} = (0,0+2] = (\beta(b@y),\beta(b@y) + \sigma(y,z)]\text{.}
\end{align*}

The next definition establishes a compatibility relation between path-based arrival time mappings,
analogous to Definition~\ref{def:order:compatible}.
\begin{definition}\label{def:amap:compatible}
  Two path-based arrival time mappings $\alpha_1$ and $\alpha_2$ for the event set $\events$
  for a weighted MAPF problem
  are \emph{compatible} if
  \begin{enumerate}
  \item $\alpha_1(a@u) < \alpha_1(a@v)$ iff $\alpha_2(a@u) < \alpha_2(a@v)$ for all $a@u,a@v \in \events$, and
  \item $\alpha_1(a@u) < \alpha_1(b@u)$ iff $\alpha_2(a@u) < \alpha_2(b@u)$ for all $a@u,b@u \in \events$.
  \end{enumerate}
\end{definition}

Given two arrival time mappings $\alpha_1$ and $\alpha_2$ for event set~$\events$,
we say that $\alpha_1$ is \emph{smaller} than $\alpha_2$
if $\alpha_1(\epsilon) \leq \alpha_2(\epsilon)$ for all $\epsilon \in \events$.
Similar to Section~\ref{sec:ordering}, this allows us to distinguish minimal arrival time mappings below.
\note{R:
The idea of the minimal compatible mapping is that it characterizes solutions computed by \clingodl.
The common basis here is vertex conflict-freeness, we can optionally throw in swap and $\sigma$-follow conflict-freeness.
This needs a detailed proof to ensure no cases are omitted.}
\note{T: Seems we are not using it otherwise, shall we expand your thought\dots?}
\begin{proposition}\label{stm:amap:onto}
  For each conflict-free path-based arrival time mapping for a weighted MAPF problem,
  there is a unique minimally compatible conflict-free path-based arrival time mapping for the problem.
\end{proposition}

Next, we formalize the correspondence between arrival time mappings and path-based plans.
Unlike in Proposition~\ref{stm:plan:onto:order},
arrival time mappings not just induce partial orders but total preorders among events.
This allows us to establish a stronger relationship between mappings and plans.
To this end,
we use the following auxiliary concept:
A plan of length $n$ has \textit{excess length}, if $\pi_a(i)=g_a$ at some position $0 \leq i < n$ for all agents $a\in A$ .
\begin{proposition}\label{stm:wplan:amap}
  Given a weighted MAPF problem~$(V,E,A,\delta,\sigma)$,
  there is a one-to-one correspondence between
  path-based plans for $(V,E,A,\delta,\sigma)$ not having excess length
  and
  conflict-free path-based arrival time mappings for $(V,E,A,\delta,\sigma)$.
\end{proposition}
Note that we can map any path-based plan to a plan as required in the proposition
by reducing the length of the plan to the latest arrival time of any agent at its goal vertex.

With this correspondence between plans and mappings,
we now identify concise conditions to compactly encode weighted MAPF problems
using ASP augmented with difference constraints in Section~\ref{sec:wmapf:encodings:sequence}.

We begin with vertex and $\sigma$-follow conflict-free mappings.
\begin{proposition}\label{stm:amap:vfconflict}
  A path-based arrival time mapping~$\alpha$ for the event set $\events$ for a weighted MAPF problem~$(V,E,A,\delta,\sigma)$
  is vertex and $\sigma$-follow conflict-free,
  if for all $a@u,b@u \in \events$ with $a \neq b$ we either have that
  \begin{enumerate}
  \item there is an event~$a@v \in \events$ such that $a@u \precdot_{\alpha_a} a@v$ and $\beta(a@u) + \sigma(u,v) < \alpha(b@u)$, or
  \item there is an event~$b@v \in \events$ such that $b@u \precdot_{\alpha_b} b@v$ and $\beta(b@u) + \sigma(u,v) < \alpha(a@u)$.
  \end{enumerate}
\end{proposition}
Note that this proposition has a very intuitive reading:
Whenever there is a potential conflict between two agents at a vertex,
one of them has to move out of the way first,
while respecting the safety period given by $\sigma$.

Finally, we show that we can prevent swap conflicts by selecting the right cases in the above proposition.
Considering a path-based arrival time mapping and the first case of Proposition~\ref{stm:amap:vfconflict}
(along with all involved entities),
we have
\begin{align*}
\sigma(u,v) &\geq 0\text{,}\\
\alpha(a@u) + \delta(u,v) &\leq \alpha(a@v)\text{,}\\
\beta(a@u) & = \alpha(a@v) - \delta(u,v)\text{, and}\\
\beta(a@u) + \sigma(u,v) &< \alpha(b@u)\text{.}
\intertext{We conclude}
\alpha(a@u) &\leq \alpha(a@v) - \delta(u,v)\\
& \leq \alpha(a@v) - \delta(u,v) + \sigma(u,v)\\
& < \alpha(b@u)\text{.}
\end{align*}
Since the cases are symmetric,
we obtain $\alpha(a@u) < \alpha(b@u)$ in the first and $\alpha(b@u) < \alpha(a@u)$ in the second case.
Using the picture
\begin{center}
\begin{tikzpicture}
\node (in) at (-1,0) {};
\node[draw,circle] (u) at (0,0) {$u$};
\node[draw,circle] (v) at (1,0) {$v$};
\node (out) at (2,0) {};
\draw[->] (in) -> node[above left] {$b$} (u);
\draw[->] (out) -> node[above right] {$a$} (v);
\draw[->] (u) .. controls (0.5,-0.15) .. (v);
\draw[->] (v) .. controls (0.5,+0.15) .. (u);
\end{tikzpicture}
\end{center}
as a guide,
we observe that we can avoid the swap conflicts for agents~$a$ and~$b$ moving along edges~$(v,u)$ and~$(u,v)$
in a vertex and $\sigma$-follow conflict-free path-based plan
if we ensure that either $\alpha(a@u) < \alpha(b@u)$ or $\alpha(b@v) < \alpha(a@v)$.
The above derivation shows that,
for a vertex and $\sigma$-follow conflict-free path-based plan,
$\alpha(a@u) < \alpha(b@u)$ is implied by $\beta(a@u) + \sigma(u,v) < \alpha(b@u)$
and $\alpha(b@v) < \alpha(a@v)$ by $\beta(b@v) + \sigma(v,u) < \alpha(a@v)$
providing the basis for the following proposition:
\begin{proposition}\label{stm:amap:sconflict}
  A vertex and $\sigma$-follow conflict-free path-based arrival time mapping~$\alpha$ for a weighted MAPF problem~$(V,E,A,\delta,\sigma)$ is swap conflict-free,
  if we either have
  \begin{enumerate}
  \item $\beta(a@u) + \sigma(u,v) < \alpha(b@u)$, or
  \item $\beta(b@v) + \sigma(v,u) < \alpha(a@v)$.
  \end{enumerate}
  for all $a@u \precdot_{\alpha_a} a@v$ and $b@v \precdot_{\alpha_b} b@u$ with $a \neq b$.
\end{proposition}

 \subsection{Fact format}\label{sec:wmapf:format}

We represent a weighted MAPF problem $(V,E,A,\delta,\sigma)$
as a set of facts $\Gamma(V,E,A,\delta)$
consisting of the facts in $\Gamma(V,E,A)$ and
atoms \ASP|edge($u$,$v$,$\delta(u,v)$)| for each edge $(u,v)\in E$.
For $\sigma$ we distinguish the three special cases defined at the beginning of this section, namely,
$\sigma_{\mathrm{v}}$, $\sigma_{\mathrm{e}}$, and $\sigma_{\mathrm{s},d}$ for $d \in \mathbb{N}_0$.
Their selection is controlled via parameters $m \in \{\text{\ASP|vf|},\text{\ASP|ef|},\text{\ASP|sf|}\}$ and $d\in \mathbb{N}_0$.
In the following, we use $\sigma_{m,d}$ to refer to the chosen distance function based on the given parameters.

As in Section~\ref{sec:mapf:format},
we represent plans by predicates
\ASP|move/3| or \ASP|move/4|.
A path or stroll $\pi$ of an agent $a$ is then represented by atoms of form
\ASP|move($a$,$\pi(i)$,$\pi(j)$)| or \ASP|move($a$,$\pi(i)$,$\pi(j)$,$i+1$)|,
each representing a move $(i,j) \in \mathcal{M}_{\pi}$.
We use the variant with time points in the next section
and drop them once more in Section~\ref{sec:wmapf:encodings:sequence}.

\subsection{Vanilla Encoding for Weighted MAPF}\label{sec:wmapf:encodings:vanilla}

The encoding in Listing~\ref{lst:wmapf:vanilla} is used to find conflict-free plans~$\{\pi_a\}_{a\in A}$ of length $n$
for weighted MAPF problems~$(V,E,A,\delta,\sigma_{m,d})$
parametrized by $m$ and $d$.
It refines the encoding in Listing~\ref{lst:mapf:vanilla} for dealing with durative move actions.
\begin{lstlisting}[language=clingoht,
                   caption={Vanilla Encoding for Weighted MAPF.},
                   label={lst:wmapf:vanilla}]
block_edge(U,V,S) :- edge(U,V,D), edge(V,U,E), S=D-E..D-1.|\label{lst:wmapf:vanilla:isc}|

block_vertex(U,V,S) :- edge(U,V,D), S=0..D-1, $m$=vf.|\label{lst:wmapf:vanilla:vf}|
block_vertex(U,V,S) :- edge(U,V,D), S=0..D-2, $m$=ef.|\label{lst:wmapf:vanilla:ef}|
block_vertex(U,V,S) :- edge(U,V,D), S=0..|\LC{d}|-1, $m$=sf.|\label{lst:wmapf:vanilla:sf}|

{ move(A,U,V,T): edge(U,V,_) } <= 1 :- agent(A), T=1..|\LC{n}|.|\label{lst:wmapf:vanilla:move}|

at(A,U,0) :- start(A,U).|\label{lst:wmapf:vanilla:init}|
at(A,V,T) :- move(A,U,V,T-D+1), edge(U,V,D), T=1..|\LC{n}|.|\label{lst:wmapf:vanilla:effect}|
at(A,U,T) :- at(A,U,T-1), not move(A,U,_,T), T=1..|\LC{n}|.|\label{lst:wmapf:vanilla:inertia}|

:- move(A,U,_,T), not at(A,U,T-1).|\label{lst:wmapf:vanilla:valid}|
:- goal(A,U), not at(A,U,|\LC{n}|).|\label{lst:wmapf:vanilla:goal}|

:- { at(A,U,T) } > 1, vertex(U), T=0..|\LC{n}|.|\label{lst:wmapf:vanilla:vconflict}|
:- move(_,U,V,T), move(_,V,U,T+S), block_edge(U,V,S).|\label{lst:wmapf:vanilla:sconflict}|
:- at(A,U,T), move(B,U,V,T-S), A!=B, block_vertex(U,V,S).|\label{lst:wmapf:vanilla:fconflict}|

location(A,U,T)     :- at(A,U,T).|\label{lst:wmapf:vanilla:loc:vertex}|
location(A,(U,V),T) :- move(A,U,V,T-S), edge(U,V,D), S=0..D-2, T=1..|\LC{n}|.|\label{lst:wmapf:vanilla:loc:edge}|
:- { location(A,U,T) } != 1, agent(A), T=1..|\LC{n}|.|\label{lst:wmapf:vanilla:loc:unique}|
\end{lstlisting}

We define \ASP|block_edge($u$,$v$,$s$)| in Line~\ref{lst:wmapf:vanilla:isc} with~$s \in [\delta(u,v)-\delta(v,u),\delta(u,v))$ for edges $(u,v)$ and $(v,u)$.
Similarly,
\ASP|block_vertex($u$,$v$,$s$)| is defined in Lines~\ref{lst:wmapf:vanilla:vf}, \ref{lst:wmapf:vanilla:ef}, and \ref{lst:wmapf:vanilla:sf}
with $s \in [0, \sigma(u,v))$ for edges $(u,v)$ based on the selected type of $\sigma$-follow conflict.
We delay a further explanation of both intervals until we discuss swap and $\sigma$-follow conflicts below.

As in Listing~\ref{lst:mapf:vanilla},
we use a choice rule in Line~\ref{lst:wmapf:vanilla:move} to generate move candidates.
For each time point $1 \leq t \leq n$ and agent $a \in A$,
we choose at most one \ASP|move($a$,$u$,$v$,$t$)| for an edge $(u,v) \in E$.
The selected atoms correspond to a set of moves of form $(t-1,t-1+\delta(u,v))$ for an agent $a$ moving from vertex $u$ to $v$ at time point $t$.
In the following,
we add further rules to ensure that this set corresponds to the set $\mathcal{M}_{\pi_a}$ of moves for each stroll $\pi_a$.

In Lines~\ref{lst:wmapf:vanilla:init}, \ref{lst:wmapf:vanilla:effect}, and \ref{lst:wmapf:vanilla:inertia},
we generate candidates for strolls $\pi_a$ from the start positions~$s_a$ of agents $a \in A$ and selected moves.
As in Listing~\ref{lst:mapf:vanilla},
Line~\ref{lst:wmapf:vanilla:init} ensures that $\pi_a(0) = s_a$ encoded by \ASP|at($a$,$s_0$,0)|.
In the next line,
we derive agent positions from \ASP|move($a$,$u$,$v$,$t-\delta(u,v)+1$)|.
Using the above correspondence, this move corresponds to the pair $(t'-1,t'-1+\delta(u,v))$ with $t'=t-\delta(u,v)+1$.
Hence, we obtain the move $(t-\delta(u,v),t)$ and derive the agent position \ASP|at($a$,$u$,$t$)| establishing $\pi_a(t) = u$.
The last line in the block encodes that an agent stays at its position if it has not been moved.
That is, if we have $\pi_a(t-1) = u$ and there is no move~$(t-1,t-1+\delta(u,v))$ for any edge $(u,v)$,
we obtain $\pi_a(t) = u$ captured by \ASP|at($a$,$u$,$t$)|.
Note that we do not explicitly represent agents located at auxiliary vertices as used in Definition~\ref{def:wstroll}.

As in  Listing~\ref{lst:mapf:vanilla},
the vertices in the stroll candidates~$\pi_a$ are at this point
not necessarily connected by edges nor do they lead to goal vertices~$g_a$.
Connectedness is ensured by the integrity constraint in Line~\ref{lst:wmapf:vanilla:valid}.
We discard candidates whenever there is no agent at the source vertex of a move.
Since a move \ASP|move($a$,$u$,$v$,$t$)| corresponds to the pair $(t-1,t-1+\delta(u,v))$,
we require $\pi_a(t-1) = u$ via \ASP|not at($a$,$u$,$t-1$)|.
Now,
stroll candidates are indeed walks in the auxiliary graph of Definition~\ref{def:wstroll}.
The integrity constraint in Line~\ref{lst:wmapf:vanilla:goal} ensures that $\pi_a(n) = g_a$,
that is, that all strolls lead to goal vertices.

At this stage,
we have a plan $\{\pi_a\}_{a\in A}$ being not necessarily conflict-free.
This is addressed in Lines~\ref{lst:wmapf:vanilla:vconflict}, \ref{lst:wmapf:vanilla:sconflict}, and~\ref{lst:wmapf:vanilla:fconflict},
where we ensure that the plan is free of vertex, swap, and $\sigma$-follow conflicts, respectively.
The first line is identical to the one in Listing~\ref{lst:mapf:vanilla}
but the integrity constraint still implements Definition~\ref{def:wplan:conflict}.\ref{def:wplan:vconflict},
discarding any plan where two agents are at the same vertex.

The second one in Line~\ref{lst:wmapf:vanilla:sconflict} ensures that there are no swap conflicts.
The atom \ASP|move(_,$u$,$v$,$t$)| represents the pair $(t-1,t-1+\delta(u,v))$.
By Definition~\ref{def:wplan:conflict}.\ref{def:wplan:sconflict},
we have to ensure that there is no move starting at time point $l$ such that $t-1+\delta(u,v) \in (l,l+\delta(v,u)]$.
Rearranging,
we obtain $l \in [t-1] + [\delta(u,v)-\delta(v,u),\delta(u,v))$.
Using $s \in [\delta(u,v)-\delta(v,u),\delta(u,v))$ as obtained via \ASP|block_edge|,
there must be no move representing the pair $(t-1+s,t-1+s+\delta(v,u))$
corresponding to \ASP|move(_,$v$,$u$,$t+s$)|.
Note that we project out agents
because an agent cannot cause a swap conflict with itself.\comment{{R}:
  With a little lemma stating that we can use the departure time as well as the arrival time in the check,
  we could avoid having to juggle with two durations.
  Let me know what you think!\par
  {R}: Note that we can only project out agents for swap but not for follow conflicts
  because we do not want to stop an agent from following itself.}
\comment{T: tough stuff}

To simplify the treatment of follow-conflicts in Line~\ref{lst:wmapf:vanilla:fconflict},
we presuppose that there are no vertex conflicts.
This allows us to compare a vertex position with a move instead of two moves.
Hence,
we slightly deviate from Definition~\ref{def:wplan:conflict}.\ref{def:wplan:fconflict} and check that there is no $t \in (k,k+\sigma(u,v)]$
for an agent~$a$ located at vertex~$u$ at time point $t$
while there is another agent~$b$ moving from $u$ to $v$ at time point $k$.
We begin with \ASP|at($a$,$u$,$t$)| indicating $\pi_a(t)=u$.
Rearranging the above check, we obtain $k \in [t-1] - [0,\sigma(u,v))$.
Using $s \in [0,\sigma(u,v))$ as obtained via \ASP|block_vertex|,
there must be no move for another agent $b$ representing the pair $(t-1-s,t-1-s+\delta(u,v))$
corresponding to \ASP|move($b$,$u$,$v$,$t-s$)|.
\comment{T: tough stuff}

At last,
we add a redundant check in Lines~\ref{lst:wmapf:vanilla:loc:vertex}, \ref{lst:wmapf:vanilla:loc:edge}, and~\ref{lst:wmapf:vanilla:loc:unique}
asserting that any agent must be at exactly one location
where a location is either a vertex or an edge.
This check is meant to improve solving performance,
analogous to Line~\ref{lst:mapf:vanilla:loc:unique} in Listing~\ref{lst:mapf:vanilla}.

Finally,
we establish the following correspondence between the stable models of our vanilla encoding along with facts representing a weighted MAPF problem
and conflict-free plans for the same weighted MAPF problem.
\begin{proposition}[Soundness]\label{prop:wmapf:vanilla:sound}
  Given a weighted MAPF problem $(V,E,A,\delta,\sigma_{m,d})$,
  let $P$ be the union of\/ $\Gamma(V,E,A,\delta)$ and the encoding in Listing~\ref{lst:wmapf:vanilla}
  for parameters $m$, $d$ and $n\in\mathbb{N}_0$.

  If $X$ is a stable model of $P$, then $\{\pi_a\}_{a\in A}$ with
\begin{align*}
  \pi_a(i) &= \begin{cases}
  v & \text{if \ASP|at($a$,$v$,$i$)|} \in X\text{, or} \\
  e^{u,v}_{i-k} & \text{if \ASP|move($a$,$u$,$v$,$k+1$)|} \in X \text{ and } k < i < k+\delta(u,v)
  \end{cases}
  \end{align*}
for $a \in A$ and $0 \leq i \leq n$ is a conflict-free plan of length $n$ for $(V,E,A,\delta,\sigma_{m,d})$.
\end{proposition}
Conversely, we can also construct stable models from plans.
\begin{proposition}[Completeness]\label{prop:wmapf:vanilla:complete}
  Given a weighted MAPF problem $(V,E,A,\delta,\sigma_{m,d})$,
  let $P$ be the union of\/ $\Gamma(V,E,A,\delta)$ and the encoding in Listing~\ref{lst:wmapf:vanilla}
  for parameters $m$, $d$ and $n\in\mathbb{N}_0$.

  If \(\{\pi_a\}_{a\in A}\) is a conflict-free plan of length $n$ for $(V,E,A,\delta,\sigma_{m,d})$,
  then
  the set
  \(
  \Gamma(V,E,A,\delta) \cup
  \{ \text{\ASP|at($a$,$v$,$i$)|} \mid a\in A \text{,\ } v \in V \text{, } i \in \iota_{\pi_a}(v) \} \cup
  \{ \text{\ASP|move($a$,$\pi_a(i)$,$\pi_a(j)$,$i+1$)|} \mid (i,j) \in \mathcal{M}_{\pi_a} \}  \)
  is a stable model of $P$
  (omitting atoms over auxiliary predicates \ASP|block_edge| and \ASP|block_vertex|).
\end{proposition}

\subsection{Sequence Encoding for Weighted MAPF}
\label{sec:wmapf:encodings:sequence}
\comment{T: Sequence Encoding/Ordering Encoding}
In analogy to Section~\ref{sec:mapf:encoding:ordering},
we now present a solution to weighted MAPF based upon arrival time mappings.
As above, our solution drops time steps and abolishes the need for limiting the length of plans.
Also, it consists of three parts:
For producing candidate paths and event orders,
we reclaim the encodings in Listing~\ref{lst:mapf:path} and~\ref{lst:mapf:order}
from Section~\ref{sec:mapf:encoding:ordering}, respectively.
However, to account for durations,
we
use difference rather than acyclicity constraints and accordingly
replace the encoding in Listing~\ref{lst:mapf:check} by the one in Listing~\ref{lst:wmapf:sequence}.

\begin{lstlisting}[language=clingoht,
                   caption={Encoding for Minimal Conflict-free Arrival Time Mappings for Weighted MAPF},
                   label={lst:wmapf:sequence}]
:- move(A,U,V), move(B,V,U), A<B,
   not resolve(A,B,V), not resolve(B,A,U).|\label{lst:mapf:order:sconflict}|

duration(U,V,1)     :- edge(U,V,_), $m$=vf.|\label{lst:wmapf:sequence:vf}|
duration(U,V,0)     :- edge(U,V,_), $m$=ef.|\label{lst:wmapf:sequence:ef}|
duration(U,V,|\LC{d}|-D+1) :- edge(U,V,D), $m$=sf.|\label{lst:wmapf:sequence:sf}|

&diff{(A,U)+D}<=(A,V) :- move(A,U,V), edge(U,V,D).|\label{lst:wmapf:sequence:total}|
&diff{(A,V)+W}<=(B,U) :- resolve(A,B,U), move(A,U,V), duration(U,V,W).|\label{lst:wmapf:sequence:conflict}|
\end{lstlisting}
More precisely,
the encoding in Listing~\ref{lst:wmapf:sequence} is used to find minimal conflict-free arrival time mappings
for weighed MAPF problems~$(V,E,A,\delta,\sigma_{m,d})$
for parameters $m$ and $d$.
The two parameters are used to select which type of $\sigma$-follow conflict to prevent.
Both take the same values as for the encoding in Listing~\ref{lst:wmapf:vanilla}.

Recall that the candidate paths produced by Listing~\ref{lst:mapf:path} are represented by atoms of form \ASP|move($a$,$u$,$v$)|,
which indicate that an agent $a$ moves along edge $(u,v)$.
The selected moves form paths leading from start to goal vertices as well as separate cycles with at least two vertices.\footnote{These superfluous cycles are pruned by the difference constraints in Listing~\ref{lst:wmapf:sequence},
  just as done by the acyclicity constraints in Listing~\ref{lst:mapf:check}.}
Also, recall that the candidate event orders produced by Listing~\ref{lst:mapf:order} are represented by atoms of form~\ASP|resolve($a$,$b$,$u$)|,
indicating that agent~$a$ has to move through vertex $u$ before~$b$ does.\comment{{R}:
 We need to argue here why this order prevents swap conflicts also in the case with durations.}

Lines~\ref{lst:wmapf:sequence:vf}, \ref{lst:wmapf:sequence:ef}, and~\ref{lst:wmapf:sequence:sf} in Listing~\ref{lst:wmapf:sequence}
precompute durations depending on the type of follow-conflict in view of Proposition~\ref{stm:amap:vfconflict}.
Note that the departure time of agent~$a$ moving from~$u$ to~$v$ can be represented as $\beta(a@u) = \alpha(a@v) - \delta(u,v)$.
Considering the first case,
we obtain $\alpha(a@v) - \delta(u,v) + \sigma(u,v) < \alpha(b@u)$.
This can equivalently be written as $\alpha(a@v) + d \leq \alpha(b@u)$ with $d=\sigma(u,v)-\delta(u,v)+1$;
the atom~\ASP|duration($u$,$v$,$d$)| captures the duration~$d$ for agents moving along an edge~$(u,v)$.
Since the second case in the proposition is symmetric,
we can use the same atom.
\comment{T: tough stuff}

In Line~\ref{lst:wmapf:sequence:total},
we ensure that there is a path-based mapping by means of difference constraints.
This is done in analogy to Line~\ref{lst:mapf:check:path} in Listing~\ref{lst:mapf:check}
by inspecting the selected path candidate via \ASP|move| atoms.
The rule with the difference constraint of form \ASP|&diff{($a$,$u$)+$d$}<=($a$,$v$)| in the head
can be read as: $\alpha(a@u)+d \leq \alpha(a@v)$ for all agents~$a$ moving along edge~$(u,v)$ with weight~$d$.
Observe that such difference constraints cannot be satisfied if our path candidate contains cycles;
for a single path, however,
a mapping can be constructed.
Hence, at this point, we have made sure that there is a path-based arrival time mapping $\alpha$ as in Definition~\ref{def:amap:path-based}.

The rule in Line~\ref{lst:wmapf:sequence:conflict}
implements Proposition~\ref{stm:amap:vfconflict}.
The mutually exclusive atoms~\ASP|resolve($a$,$b$,$u$)| and \ASP|resolve($b$,$a$,$u$)|
distinguish the two cases in Proposition~\ref{stm:amap:vfconflict}.
The candidate order guarantees
that we have exactly one of them for each potential conflict vertex~$u$ involving agents $a$ and $b$.
Let us consider the case that \ASP|resolve($a$,$b$,$u$)| is true.
Since such atoms are derived from moves,
there must be a true \ASP|move($a$,$u$,$v$)| atom
corresponding to some $a@v \in \events$ with $a@u \precdot_{\alpha_a} a@v$.
Via the difference constraint in the head,
we establish $\beta(a@u) + \sigma(u,v) < \alpha(b@u)$.\comment{T: How?}
The second case is symmetric.
At this point,
we have shown that we have a vertex and $\sigma$-follow conflict-free arrival time mapping.

Finally,
we can use Proposition~\ref{stm:amap:sconflict} to argue that the arrival time mapping is also swap conflict-free.
The constraint in Line~\ref{lst:mapf:order:sconflict} ensures that the candidate order satisfies at least one of the two cases in the proposition.\comment{T: How?}
Furthermore, the path-based nature of the arrival time mapping ensures that only one of them applies at a time.

In the remainder of this section,
we demonstrate how to obtain arrival time mappings from stable models of the above programs
and vice versa.
To begin with, we note that once difference constraints occur in a logic program,
its stable models come with an integer assignment witnessing the satisfiability of all derived difference constraints.
More precisely,
this witness assigns integers to variables occurring in \ASP|&diff| head atoms
whose corresponding bodies are satisfied by the stable model;
all other such variables remain undefined.
The witnessing assignment satisfies all constraints corresponding to derived \ASP|&diff| atoms.
In the encoding in Listing~\ref{lst:wmapf:sequence},
witnesses assign integers $i$ to variables in the form of pairs $(a,u)$ of agents and vertices;
we represent such witnesses as sets of elements of form $(a,u)=i$.
Also, we implicitly add the following rule to Listing~\ref{lst:wmapf:sequence}:\footnote{For example,
  without the additional rule,
  a weighted MAPF instance with just one agent and equal start and goal vertex would have an empty assignment as witness.
  With the additional rule, the assignment maps the corresponding event to zero.}
\begin{lstlisting}[language=clingoht,float=false,frame=,numbers=none]
&diff{(A,U)}<=(A,V) :- start(A,U), goal(A,V).
\end{lstlisting}
Furthermore,
\clingodl\ computes a canonical witness by
assigning variables the smallest possible integer greater or equal to zero.\comment{{R2TP}:
  Parts of this, might be something for the introduction of this section already.\par
  T: relate difference constraints with \clingodl\ somewhere}
The following propositions state that these assignments correspond to minimal arrival time mappings.
\begin{proposition}[Soundness]\label{stm:wmapf:model-to-mapping}
  Let $(V,E,A,\delta,\sigma_{m,d})$ be a weighted MAPF problem and $P$ be the union of\/ $\Gamma(V,E,A,\delta)$ and
  the encodings in Listings~\ref{lst:mapf:path}, \ref{lst:mapf:order}, and~\ref{lst:wmapf:sequence}
  with parameters $m$ and $d$.

  If $X$ is stable model of $P$ with witness $W$,
  then the mapping $\{ a@u \mapsto i \mid (a,u)=i \in W \}$ is
  a minimal conflict-free path-based arrival time mapping for $(V,E,A,\delta,\sigma_{m,d})$.
\end{proposition}
Together with Proposition~\ref{stm:wplan:amap},
this implies the existence of a corresponding path-based plan,
which can be constructed as described in Section~\ref{sec:mapping}.
\begin{proposition}[Completeness]\label{stm:wmapf:mapping-to-model}
  Let $(V,E,A,\delta,\sigma_{m,d})$ be a weighted MAPF problem and $P$ be the union of\/ $\Gamma(V,E,A,\delta)$ and
  the encodings in Listings~\ref{lst:mapf:path}, \ref{lst:mapf:order}, and~\ref{lst:wmapf:sequence}
  with parameters $m$ and $d$.

  If $\alpha$ is a minimal conflict-free path-based arrival time mapping over event set $\events$ for $(V,E,A,\delta,\sigma_{m,d})$,
  then
  the smallest set $X$ such that
  \begin{enumerate}
  \item $\Gamma(V,E,A,\delta) \subseteq X$,
  \item $\text{\ASP|move($a$,$u$,$v$)|} \in X$ for all $a@u \precdot_{\alpha_a} a@v$, and
  \item $\text{\ASP|resolve($a$,$b$,$u$)|} \in X$ for all $a@v \prec b@u$ and $a@u \precdot_{\alpha_a} a@v$ with $a \neq b$
  \item []  (omitting auxiliary atoms over predicate \ASP|duration|)
  \end{enumerate}
  is a stable model of $P$ with witness $\{ (a,u)=\alpha(a@u) \mid a@u \in \events \}$.
\end{proposition}
This and Proposition~\ref{stm:wplan:amap} implies that
each path-based plan for a weighted MAPF problem
has a corresponding stable model and witness of the weighted MAPF encodings.

 \section{Experiments}\label{sec:experiments}
To evaluate our encoding variants,
we built an ASP-based benchmark generator for different types of grid-based MAPF problems,\footnote{\url{https://github.com/krr-up/mapf-instance-generator}}
viz.\ random and room configurations.
The instances are created by choosing subgraphs of a square grid graph
$(V,E)$ with
$V=\{v_{ij} \mid 1 \leq i,j \leq n\}$ and
$E=\{ (v_{ij},v_{kl}) \in V \times V \mid |i-k|+|j-l|=1 \}$ for some $n>0$.
The generated graphs are undirected and connected.
Agents' start and goal vertices are picked randomly.
For the maze instances, the graph forms a tree, that is, there is exactly one path between two vertices.
For room instances, the grid is evenly divided into rooms separated by walls of width one;
neighboring rooms are connected by choosing up to one random vertex as a door.
For random ones, a certain percentage of the vertices of the square grid graph is chosen.

Instances are built via the following parameters:
$10 \leq n \leq 40$ for the size of the underlying square grid graph,
between $5$ and $30$ agents,
square rooms of size $1$ to $5$, and
vertices are chosen with a 50\% probability for the random instances.
In total, we consider 199 instances;
94 (24 maze, 56 random, 25 room) of them have at least one conflict-free plan,
and 105 (56 random, 38 room) have at least one conflict-free path-based plan.
In total, we consider 105 instances (56 random, 38 room) with at least one conflict-free path-based plan.
We ran all instances on a compute cluster with Intel Xeon E5-2650v4@2.9GHz~CPUs with 64GB of memory running
Debian Linux~10.\footnote{\url{https://www.cs.uni-potsdam.de/bs/research/labs.html\#hardware}}
We used a timeout of 3h and limited the memory to 16GB per instance.

We consider the vanilla encoding (\emph{vanilla}) in Listing~\ref{lst:mapf:vanilla} and
the event ordering encoding (\emph{order AC}) described in Section~\ref{sec:mapf:encoding:ordering}.
Furthermore, we use a variant of the latter replacing the acyclicity constraints in Listing~\ref{lst:mapf:check}
with difference constraints (\emph{order DC}) and \clingodl\ as solver:
\begin{lstlisting}[language=clingoht,float=false,frame=]
&diff{(A,U)+1}<=(A,V) :- move(A,U,V).|\label{lst:mapf:sequence:total}|
&diff{(A,V)+1}<=(B,U) :- resolve(A,B,U), move(A,U,V).|\label{lst:mapf:sequence:conflict}|
\end{lstlisting}
Finally, we use a plain ASP encoding (\emph{order ASP}) to check if the generated order is acyclic.
Without going into details, we mention that this check results in groundings of cubic size in the number of possible events.

\comment{T: rewrite in view of material in refinements}
To be able to use the event ordering encodings also on problems that do not have path-based plans,
we generalize the above encodings to allow agents to stop at intermediate waypoints (\emph{order \{AC,DC,ASP\}+WP}).
The encoding generates $w\geq0$ waypoints for agents using a choice rule.
Given a MAPF problem,
we create $w+1$ instances of the event ordering encodings described above adjusting start and goal positions.
Considering the start and the goal positions of the MAPF problem as first and last waypoint;
the combined encoding finds path-based plans between adjacent waypoints.
The concatenation of these plans is a plan for the MAPF problem.
In our setting, we only consider one waypoint.

We try to find both conflict-free (\emph{vsfc-free}) and vertex and swap conflict-free plans (\emph{vsc-free}).
In the second setting,
this means setting the mode $m$ to \ASP{fc} for the vanilla encoding.
For the DC encoding,
this means changing the difference constraint in Line~\ref{lst:mapf:sequence:total} above to use \ASP{+0} instead of \ASP{+1}
and an additional integrity constraint to discard swap conflicts.
Note that the integrity constraint does not increase the asymptotic size complexity of the grounding.
For the ASP and AC encodings,
the second setting involves a more complicated construction,
which we do not detail here.
Note that even for the AC encodings,
this involves a lot more overhead comparable to that of the order ASP setting.

We formulate the following hypotheses for evaluating the generated benchmark on the configurations described above.
The vanilla encoding should perform well for small plan lengths (\emph{H1}).
The order ASP configuration is not applicable in practice because of the too large grounding (\emph{H2}).
The order AC configuration works well for path-based plans in the vsfc-free setting even with large plan lengths (\emph{H3})
but does not work well in the vsc-free setting (\emph{H4}).
The order DC configuration works well for path-based plans with large plan lengths independent of the conflict handling (\emph{H5}).
The hypotheses for waypoint based event ordering builds on the above ones.
It should work well for large plan lengths that need extra room for agents to evade each other (\emph{H6}).

\begin{figure}
\figrule
\includegraphics{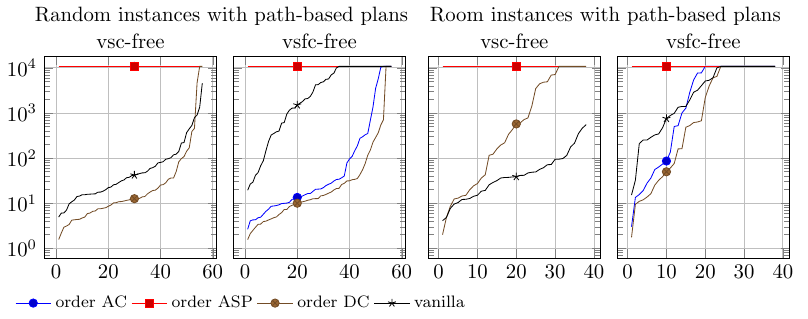}
\caption{Evaluation of random and room instances with path-based plans}\label{fig:path-based:cactus}
\figrule
\end{figure}
The plots in Figure~\ref{fig:path-based:cactus} present cactus plots showing the time (x-axis) to solve an instance (y-axis);
run times are sorted in ascending order for each curve matching one of the configurations discussed above. Here we consider only instances that have at least one path-based plan restricting the set to random and room instances.
The upper rows of plots gives results for vsc-free and the lower row vsfc-free plans.
Note that we fix the plan length for the vanilla encoding in the vsc-free setting to the optimal length such that an instance is satisfiable.
Since we did not have this information available for vsfc-free (and it is hard to compute),
we added 20\% to the plan length.

We begin with evaluating the vsc-free setting.
In line with H1,
we observe that the vanilla encoding performs well
noting that agents can move to their goal vertices rather directly.
Agreeing with H5, order DC performs well too.
On room instances it is better than vanilla,
which we attribute to the longer plan length necessary to avoid obstacles.
Interestingly, the vanilla encoding seems to work very well if follow-conflicts are permitted.
This is followed by the order DC and order DC+WP configurations agreeing with H5.
As stipulated in H2 and H4, the order ASP and order AC configurations do not work at all.
Also the settings with waypoints do not work here because they introduce too much (unnecessary) overhead.

In the vsfc-free setting,
we evaluate the random and room instances separately.
We begin with the random instances.
Note that order DC performs almost the same as in the vsc-free setting.
However, the vanilla configuration becomes much slower if plans have to be follow conflict-free.
While this is partly associated with larger plan lengths, we cannot fully explain this behavior.
Finally, we confirm H3, noting that order AC performs well here.
However, despite the less expressive AC constraints, it does not perform as well as order {DC}.
This is due to \clingodl\ implementing stronger propagation enabled using option \lstinline|--propagate=full|.
Again, the waypoint based configurations do not perform well.
The room instances are much harder in the vsfc-free setting.
Not just vanilla but also the event order based configurations are slowed down.
We attribute this to short plan length due to H1
and large overhead due to many reachable vertices for the event order settings.

\begin{figure}
\figrule
\includegraphics[width=\textwidth]{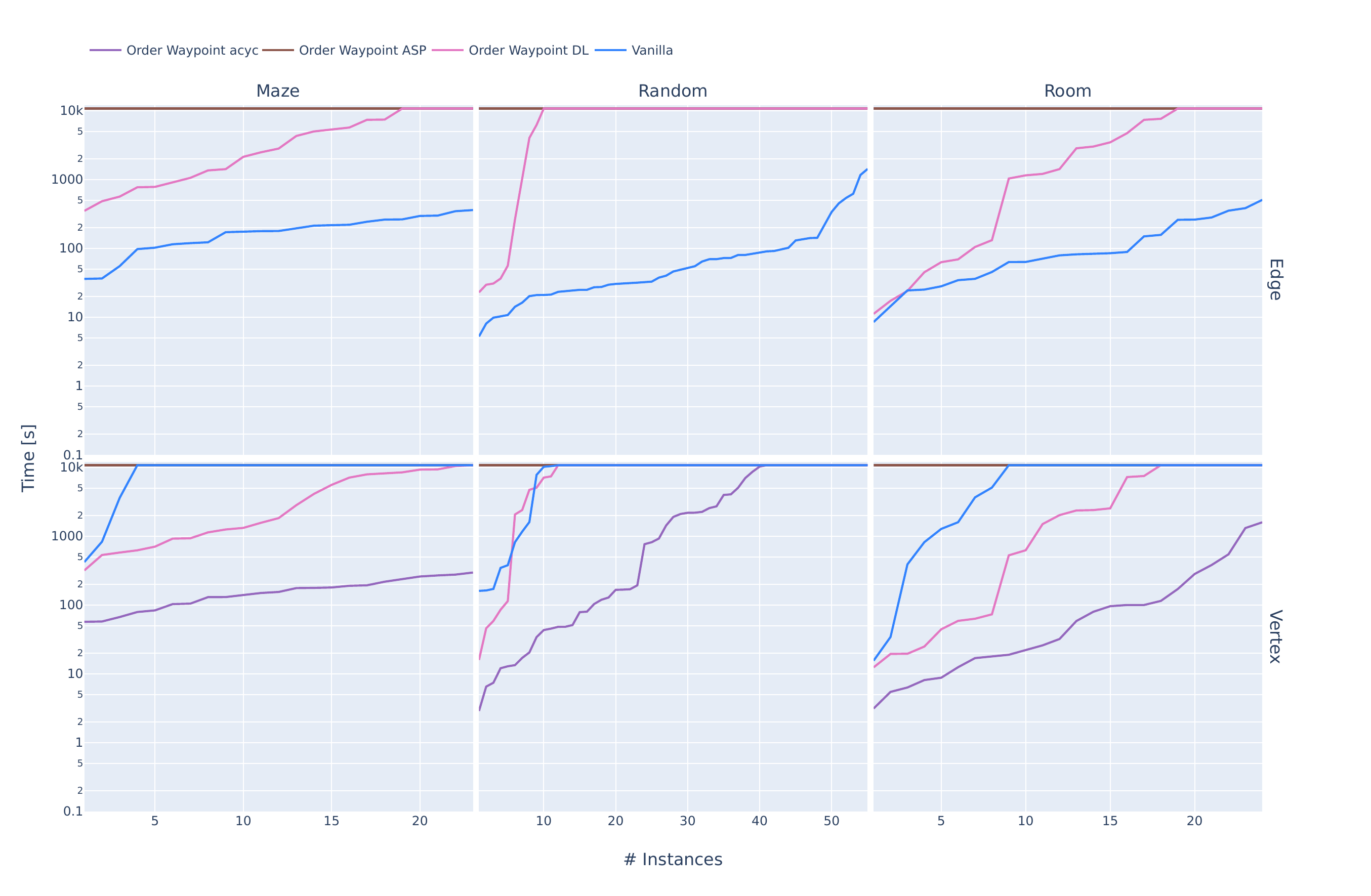}
\caption{time not path-based}\label{fig:not-path-based:cactus}
\figrule
\end{figure}
The plots in Figure~\ref{fig:not-path-based:cactus} present cactus plots for instances that do not have path-based plans.
Hence, we only consider the vanilla configuration and configurations with waypoints.
The upper row considers the vsc-free and the lower row the vsfc-free setting.
In the upper row, we obtain very similar results as for the plots in Figure~\ref{fig:path-based:cactus}.
We can confirm H6.
The previous observations carry over to the lower row.
Again the order AC configuration performs best.
We note how badly the vanilla encodings copes with follow conflicts.

 \section{Discussion}\label{sec:discussion}

As mentioned, we have already successfully applied this technique in industrial applications involving routing
and scheduling~\cite{abjoossctowa21a,hamunescwa23a,rascwachliso23a}.
This paper aims at providing an introduction to the underlying encoding techniques and their formal foundations.

Our approach builds on the characterization of plans in terms of partially ordered event sets.
This is similar to partial order planning, where a partial order is maintained among actions~\cite{sacerdoti75a}.
In MAPF,
a related idea was implemented using activity constraints from constraint-based scheduling~\cite{basvvl18a}.
Also, simple temporal networks were used to add schedules to pre-computed plans~\cite{hokucomaxuayko16a}.
Similarly, in robotics,
action dependency graphs were used to avoid online collisions by adding temporal dependencies to existing plans~\cite{bedupake20a}.
Finally,
it is worth mentioning that it was recently shown that
whenever each agent has a predefined path,
the problem of deciding if there is a MAPF solution (without any bounds on any cost function)
is NP-Hard~\cite{abgehaug23a}.

\bibliographystyle{./include/latex-class-tlp/acmtrans}

\end{document}